\documentclass[1p]{elsarticle}
\usepackage{hyperref}
\usepackage{xcolor}
\usepackage{tabularx} 
\definecolor{darkblue}{rgb}{0, 0, 0.5}
\usepackage{booktabs}
\usepackage[font={small}]{caption}
\usepackage{multirow}
\usepackage[utf8]{inputenc}
\usepackage[T2A]{fontenc} 
\usepackage{adjustbox}
\usepackage{graphicx}

\biboptions{authoryear}

\usepackage{listings}
\usepackage{color}

\definecolor{dkgreen}{rgb}{0,0.6,0}
\definecolor{gray}{rgb}{0.5,0.5,0.5}
\definecolor{mauve}{rgb}{0.58,0,0.82}
\definecolor{gray}{rgb}{0.4,0.4,0.4}
\definecolor{darkblue}{rgb}{0.0,0.0,0.6}
\definecolor{lightblue}{rgb}{0.0,0.0,0.9}
\definecolor{cyan}{rgb}{0.0,0.6,0.6}
\definecolor{darkred}{rgb}{0.6,0.0,0.0}
\definecolor{lightgray}{rgb}{0.97,0.97,0.97}

\lstdefinelanguage{XML}
{
  morestring=[s][\color{mauve}]{"}{"},
  morestring=[s][\color{black}]{>}{<},
  morecomment=[s]{<?}{?>},
  morecomment=[s][\color{dkgreen}]{<!--}{-->},
  stringstyle=\color{black},
  identifierstyle=\color{lightblue},
  keywordstyle=\color{red},
  morekeywords={xmlns,xsi,noNamespaceSchemaLocation,type,id,x,y,source,target,version,tool,transRef,roleRef,objective,eventually}
}

\begin{document}

\begin{frontmatter}

\title{On the Role of Morphological Information for Contextual Lemmatization}

\author[mymainaddress]{Olia Toporkov}
\ead{olia.toporkov@ehu.eus}

\author[mymainaddress]{Rodrigo Agerri\corref{mycorrespondingauthor}}
\ead{rodrigo.agerri@ehu.eus}

\cortext[mycorrespondingauthor]{Corresponding author}

\address[mymainaddress]{HiTZ Center - Ixa, University of the Basque Country UPV/EHU, Donostia-San Sebasti\'an, Spain}

\begin{abstract}
Lemmatization is a natural language processing (NLP) task which consists of
producing, from a given inflected word, its canonical form or lemma.
Lemmatization is one of the basic tasks that facilitate downstream NLP
applications, and is of particular importance for high-inflected languages.
Given that the process to obtain a lemma from an inflected word can be
explained by looking at its morphosyntactic category, including fine-grained
morphosyntactic information to train contextual lemmatizers has become common
practice, without considering whether that is the optimum in terms of
downstream performance. In order to address this issue, in this paper we
empirically investigate the role of morphological information to develop
contextual lemmatizers in six languages within a varied spectrum of
morphological complexity: Basque, Turkish, Russian, Czech, Spanish and English.
Furthermore, and unlike the vast majority of previous work, we also evaluate
lemmatizers in out-of-domain settings, which constitutes, after all, their most
common application use. The results of our study are rather surprising. It
turns out that providing lemmatizers with fine-grained morphological features during training
is not that beneficial, not even for agglutinative languages. In fact, modern
contextual word representations seem to implicitly encode enough morphological
information to obtain competitive contextual lemmatizers without seeing any
explicit morphological signal. Moreover, our experiments suggest that the best
lemmatizers out-of-domain are those using simple UPOS tags or those trained
without morphology and, finally, that current evaluation practices for
lemmatization are not adequate to clearly discriminate between models.
\end{abstract}


\end{frontmatter}


\section{Introduction}\label{sec:introduction}
Lemmatization is one of the basic NLP tasks which consists of converting an
inflected word form (e.g., \emph{eating, ate, eaten}) into its canonical form
(e.g., \emph{eat}), usually known as the lemma. Thus, we follow the formulation
of lemmatization as defined by the SIGMORPHON 2019 shared task
\citep{aiken-etal-2019-sigmorphon}. Lemmatization is commonly used when
performing many NLP tasks such as information retrieval, named entity
recognition, sentiment analysis, word sense disambiguation, and others. For
example, for morphologically rich languages named entities are often inflected, which
means that lemmatization is required as an additional process. Thus,
lemmatization is more challenging for languages with rich inflection as the
number of variations for every different word form in such languages is very
high. Table \ref{tab:example_inflection}  illustrates this point by showing
the differences in inflections of the word `cat' for four languages with
different morphological structure. This language sample offers a spectrum of varied complexity,
ranging from the more complex ones, Basque and Russian, to the less inflected
ones, such as Spanish and English, in that order. 

  \begin{table}[hbt!]
     \centering
    \newcolumntype{Y}{>{\centering\arraybackslash}X}
    \begin{tabularx}{\textwidth}{YYYY}
        \toprule
        {English} & 
        {Spanish} & 
        {Russian} & 
        {Basque} \\ \midrule
        cat & gato & кот & katu \\
        cats & gata & коты & katuak \\
        & gatos & кота & katua \\
        & gatas & коту & katuari \\
        & & котом & katuarekin \\
        & & коте &  katuek \\
        & & котов & katuekin \\
        & & котам & katuei \\
        & & котами & katuen \\
        & & котах & katurik \\ 
        & & & katuarentzat \\
        & & & katuentzat \\ \bottomrule
    \end{tabularx}
    \caption{Examples of inflected forms of the word `cat' in Basque, English, Spanish and Russian.}
    \label{tab:example_inflection}
    \end{table}

As we can see in Table \ref{tab:example_inflection}, the word `cat' can vary in English
by changing from singular to plural. In Spanish gender (masculine/feminine)
is also marked. Things get more complicated with languages that mark case. For example,
in Russian there are six cases while for Basque there are 16, some of which can be doubly
inflected.

\begin{figure}
        \center{\includegraphics[width=39em]
        {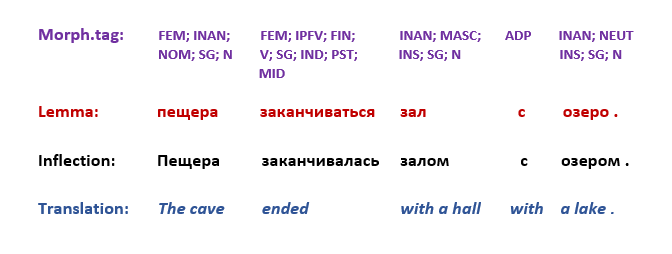}}
        \caption{Example of a morphologically tagged and lemmatized sentence in Russian using
        the UniMorph annotation scheme.}
        \label{fig:example_inflection_ru}
\end{figure}

Both the context in which it occurs and the morphosyntactic form of a word play
a crucial role to approach automatic lemmatization
\citep{mccarthy-etal-2019-sigmorphon}. Thus, in Figure
\ref{fig:example_inflection_ru} we can see a fragment of a Russian sentence
in which each inflected word form has a
corresponding lemma (in red). Furthermore, each inflected form has an
associated number of morphosyntactic features (expressed as tags) depending on its
case, number, gender, animacy and others. Morphological analysis is crucial for
lemmatization as it explains the process required to produce the lemma from the
word form, which is why it has traditionally been used as a stepping
stone to design systems to perform lemmatization. 

As many other tasks in NLP, the first approaches to lemmatization were rule-based,
but nowadays the best performing models address lemmatization as a supervised
task in which learning in context is crucial. Regardless of the learning method
used, three main trends can be observed in current contextual lemmatization:
(i) those that use gold standard or learned morphological tags to generate
features to learn lemmatization in a pipeline approach
\citep{chrupala-etal-2008-learning,yildiz-tantug-2019-morpheus}; (ii) those
that aim to jointly learn morphological tagging and lemmatization as a single
task \citep{muller-etal-2015-joint,malaviya-etal-2019-simple,straka-etal-2019-udpipe}; (iii) systems
that do not use any explicit morphological signal to learn to lemmatize
\citep{chakrabarty-etal-2017-context,bergmanis-goldwater-2018-context}. 

Research on contextual (mostly neural) lemmatization was greatly accelerated by
the first release of the Universal Dependencies (UD) data
\citep{de-marneffe-etal-2014-universal,nivre-etal-2017-universal}, but
specially by the contextual lemmatization shared task organized at SIGMORPHON
2019, which included UniMorph datasets for more than 50 languages
\citep{mccarthy-etal-2019-sigmorphon}. It should be noted that the best models in the task used
morphological information either as features \citep{yildiz-tantug-2019-morpheus}
or as part of a joint or a multitask approach \citep{straka-etal-2019-udpipe}. 
However, the large majority of previous approaches have used all the
morphological tags from UniMorph/UD assuming that fine-grained morphological
information must be always beneficial for lemmatization, especially for highly
inflected languages, but without analyzing whether that is the optimum in terms of downstream performance.

In order to address this issue, in this paper we
empirically investigate the role of morphological information to develop
contextual lemmatizers in six languages within a varied spectrum of
morphological complexity: Basque, Turkish, Russian, Czech, Spanish and English.
Furthermore, previous work has shown that morphological taggers substantially
degrade when evaluated out-of-domain, be that any type of text different from
the data used for training in terms of topic, text genre, temporality, etc.
\citep{Manning2011PartofSpeechTF}. This point led us to research whether
lemmatizers based on fine-grained morphological information will degrade more
when used out-of-domain than those requiring only coarse-grained UPOS tags. We
believe that this is also an important point because lemmatizers are mostly
used out-of-domain, namely, to lemmatize data from a different distribution
with respect to the one that was employed for training. 

Taking these issues into consideration, in this paper we set to
investigate the following research questions with respect to the actual role of
morphological information to perform contextual lemmatization. First, is
fine-grained morphological information really necessary, even for
high-inflected languages? Second, are modern context-based word representations
enough to learn competitive contextual lemmatizers without including any
explicit morphological signal for training? Third, do morphologically enriched
lemmatizers perform worse out-of-domain as the complexity of the morphological
features increases? Four, what is the optimal strategy to obtain robust
contextual lemmatizers for out-of-domain settings? Finally, are current evaluation
practices adequate to meaningfully evaluate and compare contextual
lemmatization techniques?

The conclusions from our experimental study are the following: (i) fine-grained
morphological features do not always benefit, not even for agglutinative
languages; (ii) modern contextual word representations seem to implicitly
encode enough morphological information to obtain state-of-the-art contextual
lemmatizers without seeing any explicit morphological signal; (iii) the best
lemmatizers out-of-domain are those using simple UPOS tags or those trained
without explicit morphology; (iv) current evaluation practices for
lemmatization are not adequate to clearly discriminate between models, and
other evaluation metrics are required to better understand and manifest the
shortcomings of current lemmatization techniques. The generated code and
datasets are publicly available to facilitate the reproducibility of the results and
further research on this
topic.\footnote{\url{https://github.com/oltoporkov/morphological-information-datasets}}

The rest of the paper is structured as follows. The next section discusses 
the most relevant work related to contextual lemmatization. The systems and datasets used
in our experiments are presented in Sections \ref{sec:systems} and
\ref{sec:languages_and_datasets}, respectively. Section \ref{sec:methodology}
presents the experimental setup applied to obtain the results, which are
reported in Section \ref{sec:results}. Section \ref{sec:discussion} provides a
discussion and error analysis of the results. We finish with some concluding
remarks in Section \ref{sec:conclusion}.

\section{Background}\label{sec:background}

First approaches to lemmatization consisted of systems based on dictionary
lookup and/or rule-based finite state machines \citep{karttunen-etal-1992-two,
oflazer-1993-two,eustagger-1996-alegria, van-den-bosch-daelemans-1999-memory,
Dhonnchadha2002ATM,Segalovich2003AFM,carreras-2004-freeLing,
stroppa-yvon-2005-analogical, jongejan-dalianis-2009-automatic}. Grammatical
rules in such systems, either hand-crafted or learned automatically by
using machine learning, were leveraged to perform lemmatization
together with the use of lexicons or morphological analyzers that returned the correct lemma. The problem of unseen
and rare words was solved by generating a set of exceptions added to the
general set of rules \citep{karttunen-etal-1992-two,oflazer-1993-two} or by using
a probabilistic approach \citep{Segalovich2003AFM}. Such systems resulted in very
language-dependent approaches, and in most of the cases they required huge linguistic
knowledge and effort, especially in the case of those languages with more
complex, high-inflected morphology.

The appearance of large annotated corpora with morphological information and
lemmas facilitated the development of machine learning methods for
lemmatization in multiple languages. One of the core projects that gathered
annotated corpora for more than 90 languages is the Universal Dependencies (UD)
initiative \citep{nivre-etal-2017-universal}. This project offers a unified
morphosyntactic annotation across languages with language-specific extensions
when necessary. Based on the UD data, the Universal Morphology (UniMorph)
project \citep{mccarthy-etal-2020-unimorph} converted the UD annotations into UniMorph, a universal tagset for
morphological annotation (based on \citet{SylakGlassman2016TheCA}), where each inflected
word form is associated with a lemma and a set of morphological features.
The current UniMorph dataset includes 118 languages, including extremely
low-resourced languages such as Quechua, Navajo and Haida. 

The assumption that context could help with unseen and ambiguous words led to
the creation of supervised contextual lemmatizers. The pioneer work on this
topic is perhaps the statistical contextual lemmatization model provided by Morfette
\citep{chrupala-etal-2008-learning}. Morfette uses a
Maximum Entropy classifier to predict morphological tags and lemmas in a
pipeline approach. Interestingly, instead of learning the lemmas themselves,
\citet{chrupala-etal-2008-learning} propose to learn automatically induced lemma
classes based on the shortest edit script (SES), which consists of the number of edits necessary to convert the
inflected word form into its lemma. Morfette has influenced many other
works on contextual lemmatization, such as the system of
\citet{gesmundo-samardzic-2012-lemmatisation}, IXA pipes
\citep{agerri-etal-2014-ixa,agerri_rigau}, Lemming
\citep{muller-etal-2015-joint} and the system of
\citet{malaviya-etal-2019-simple}. The importance of using context to learn
lemmatization is investigated in the work of
\citet{bergmanis-goldwater-2018-context}. They compare context-free and
context-sensitive versions of their neural lemmatizer Lematus and evaluate them
across 20 languages. Results show that including context substantially improves
lemmatization accuracy and it helps to better deal with the out-of-vocabulary
problem. 

The next step in the development of contextual lemmatization systems came with the 
supervised approaches based on deep learning algorithms and vector-based word
representations 
\citep{chakrabarty-etal-2017-context,dayanik-et-al-2018,bergmanis-goldwater-2018-context,malaviya-etal-2019-simple}.
The parallel development of the transformer architecture
\citep{Vaswani2017AttentionIA} and the appearance of BERT
\citep{devlin-etal-2019-bert} and other transformer-based masked language
models (MLMs) offered the possibility to significantly improve
lemmatization results. Thus, most of the participating systems in the
SIGMORPHON 2019 shared task on contextual lemmatization for 66 languages were based on MLMs
\citep{mccarthy-etal-2019-sigmorphon}. The
baseline provided by the task was based on the work of
\citet{malaviya-etal-2019-simple}, a system which performs joint
morphological tagging and lemmatization. 

To the best of our knowledge, current state-of-the-art results in contextual
lemmatization are provided by those models that achieved best results in the
SIGMORPHON 2019 shared task.  The highest overall accuracy was achieved by
UDPipe \citep{straka-etal-2019-udpipe}. Using UDPipe 2.0
\citep{straka-2018-udpipe} as a baseline, they added pre-trained contextualized
BERT and Flair embeddings as an additional input to the network. 
The overall accuracy (average across all languages) was 95.78, the best among all the participants. 

The second best result (95 overall word accuracy) in the task was obtained by the CHARLES-SAARLAND system
\citep{kondratyuk-2019-cross}. This system consists of a combination of a
shared BERT encoder and joint lemma and morphology tag decoder. The model uses
a two-stage training process, in which it first performs a multilingual
training over all treebanks, and then they execute the same process
monolingually, maintaining the previously learned multilingual weights.
Morphological tags in this case are calculated jointly and lemmas are also
represented as SES. The experiments are performed using multilingual BERT in
combination with the methods introduced by UDify
\citep{kondratyuk-straka-2019-75} for BERT fine-tuning and regularization. 

The third best result (94.76) was reported by Morpheus
\citep{yildiz-tantug-2019-morpheus}. Morpheus uses a two-level LSTM network
which gets as input the vector-based representations of words, morphological
tags and SES. Morpheus then aims to jointly output, for a given sequence, their
corresponding morphological labels and the SES representing the lemma class
which is later decoded into its lemma form.

Thus, it can be seen that a common trend in current contextual lemmatization is to use
the morphological information provided by the full UniMorph labels without
taking into consideration whether this is the optimal setting. Furthermore,
lemmatization techniques are only evaluated in-domain, resulting in extremely,
and perhaps deceptive, high results for the large majority of the 66 languages
included in the SIGMORPHON 2019 data.

\section{Languages and Datasets} \label{sec:languages_and_datasets}

In order to address the research questions formulated in the Introduction, we
selected the following six languages: Basque, Turkish,
Russian, Czech, Spanish and English. Such a choice will allow us to compare the
role of fine-grained morphological information to learn contextual
lemmatization within a range of languages of varied morphological complexity.
In this Section we briefly describe general morphological characteristics of
each language as well as the specific datasets used.

\subsection{Languages}\label{sec:languages}

Basque and Turkish are agglutinative languages with morphology mostly of the
suffixing type. Basque is a language isolate and does not belong to any
language group while Turkish is a member of the Oghuz group of the Turkic family.
These two languages have no grammatical gender, with some particular exceptions
for domestic animals, people and foreign words (Turkish) or in some colloquial
forms when the gender of the addressee is expressed for the second person
singular pronoun (Basque). Turkish and Basque have two number types (singular and
plural), and in Basque there is also the unmarked number (undefined or mugagabea). 
In both Turkish and Basque the cases are expressed by suffixation. 

Basque is an ergative-absolutive
language containing 16 cases, meaning that the grammatical case marks both the subject of an
intransitive verb and the object of a transitive verb. The verb conjugation is
also specific for this language: the majority of the verbs are formed by a
combination of a gerund form and a conjugated auxiliary verb.

Turkish has six general cases; nouns and adjectives are not
distinguished morphologically and adjectives can also be used as adverbs
without modifications or by doubling of the word. For verbs there are 9 simple
and 20 compound tenses. There is a relatively small set of core vocabulary and
the majority of Turkish words originate from applying derivative suffixes to
nouns and verbal stems.

The two Slavic languages, namely, Russian and Czech, which have a fusional morphological system, exhibit a highly
inflectional morphology and a wide number of morphological features. Russian
belongs to the East Slavic language group, while Czech is a West Slavic
language. These two languages have nominal declension which involves six main
grammatical cases for Russian and seven for Czech. Both languages distinguish
between two number (singular and plural) and three gender types (masculine,
feminine and neuter). Furthermore, the masculine gender is subdivided into
animate and inanimate. Verbs are conjugated for tense (past, present or future)
and mood.

Spanish is a Romance language that belongs to Indo-European language family. It
is a fusional language, which has a tendency to use a single inflectional
morpheme to denote multiple grammatical, syntactic or semantic features. Nouns
and adjectives in Spanish have two gender (male, female) and two number types
(singular and plural). Besides, some articles, pronouns and determiners also
possess a neuter gender. There are 3 main verb tenses (past, present
and future) and each verb has around fifty conjugated forms. Apart from that,
Spanish has 3 verboid forms (infinitive, gerund, past participle), perfective
and imperfective aspects for past, 4 moods and 3 persons.

Finally, English is a Germanic language, also part of Indo-European language family.
It has lower inflection in comparison to previously mentioned languages.
Only nouns, pronouns and verbs are inflected, while the rest of the parts of
speech are invariable. In English animate nouns have two genders (masculine or
feminine) and the third person singular pronouns distinguish three gender types:
masculine, feminine, and neuter, while for most of the nouns
there is no grammatical gender. Nouns have only a genitive case and personal
pronouns are mostly declined in subjective and objective cases. English
has a variety of auxiliary verbs that help to express the categories of mood
and aspect and participate in the formation of verb tenses.

\subsection{Datasets}\label{sec:datasets}

The datasets we used are distributed as part of the data used for the 
SIGMORPHON 2019 shared task \citep{mccarthy-etal-2019-sigmorphon}. The source
of the original datasets comes from the Universal
Dependencies (UD) project \citep{de-marneffe-etal-2014-universal}, but the
morphological annotations are converted from UD annotations to the UniMorph
schema \citep{kirov-etal-2018-unimorph} with the aim of increasing agreement
across languages. As our experiments will include both in-domain and
out-of-domain evaluations, we selected some datasets for each of the settings.

With respect to in-domain, we chose one corpus per language using the standard
train and development partitions. For Basque we used the Basque Dependency Treebank (BDT)
\citep{bdt_corpus}, which contains mainly literary and journalistic texts. The
corpus was manually annotated and then automatically converted to UD format.
For Czech we used the CAC treebank \citep{czech_cac_corpus} based on the
Czech Academic Corpus 2.0. This corpus includes mostly unabridged articles from
a wide range of media such as newspapers, magazines and transcripts of spoken
language from radio and TV programs. The corpus was annotated manually and then
converted to UD format. With respect to English we chose English Web Treebank (EWT)
\citep{silveira-etal-2014-gold}. This corpus includes different Web
sources: blogs, various media, e-mails, reviews and Yahoo! answers. In the EWT
corpus the lemmas were assigned by UD-converter and manually corrected. UPOS
tags were also converted to UD format from manual annotations. For
Russian we used GSD corpus, extracted from
Wikipedia and manually annotated by native speakers. In the case of Spanish we
selected the GSD corpus as well, consisting of texts from blogs, reviews, news and Wikipedia.
Finally, for Turkish we used ITU-METU-Sabanci Treebank
(IMST) \citep{sulubacak-etal-2016-universal}. It consists of well-edited sentences from a
wide range of domains, manually annotated and automatically converted to
UD format.

For the out-of-domain evaluation setting we picked the test sets of other
datasets included in UniMorph, different from the ones selected for in-domain
experimentation. In the case of Basque, only one corpus was available
in the Universal Dependencies project, so we used the Armiarma corpus which
consists of literary critics semi-automatically annotated using Eustagger
\citep{eustagger-1996-alegria}. For Czech and Turkish we used the PUD data --
part of the Parallel Universal Dependencies treebanks created for the CoNLL
2017 shared task \citep{zeman-etal-2017-conll}. The corpora consist of 1,000
sentences from the news domain and Wikipedia annotated for 18 languages. The
Czech language PUD data was manually annotated and then automatically converted to UD format. For
Turkish the original data was automatically converted to UD format,
but later manually reannotated \citep{turk-etal-2019-turkish}. In the
case of English we used the Georgetown University Multilayer (GUM) corpus
\citep{Zeldes2017}. This corpus presents a collection of annotated Web texts
from interviews, news, travel guides, academic writing, biographies and fiction
from such sources as Wikipedia, Wikinet and Reddit. Its lemmas were manually
annotated, while UPOS tags were converted to UD format from manual annotations.
In the case of Russian we used SynTagRus \citep{syntagrus}, which consists of 
texts from a variety of genres, such as contemporary fiction, popular science,
as well as news and journal articles from the 1960-2016 period. Its lemmas,
UPOS tags and morphological features were manually annotated in non-UD style and
then automatically converted to UD format. For Spanish we chose the AnCora
corpus \citep{taule-etal-2008-ancora}, which contains mainly texts from news.
All the elements of this corpus were converted to UD format from manual
annotations.

\section{Systems}\label{sec:systems}

In this section we present the systems that we will be applying in our
investigation. First, research on the role of fine-grained morphological information for
contextual lemmatization will be performed in-domain using the statistical
lemmatizer from the IXA pipes toolkit \citep{agerri_rigau} and Morpheus, the
third best system in the SIGMORPHON 2019 shared task. These two systems were
chosen due to several reasons: (i) both use morphological information as features to learn
lemmatization and, (ii) both systems use SES to represent automatically induced
lemma classes; and (iii), they both address contextual lemmatization as sequence
tagging.

In order to investigate whether modern contextual word representations are
enough to learn competitive lemmatizers both in- and out-of-domain, we train
baseline models using Flair \citep{akbik-etal-2018-contextual}, multilingual
MLMs mBERT and XLM-RoBERTa
\citep{devlin-etal-2019-bert,conneau-etal-2020-unsupervised} as well as
language-specific MLMs for each of the languages: BERTeus for Basque \citep{agerri-etal-2020-give}, slavicBERT 
for Czech \citep{arkhipov-etal-2019-tuning}, RoBERTa for English
\citep{liu2019roberta}, Russian ruBERT \citep{rubert-kuratov}, Spanish BETO
\citep{CaneteCFP2020} and BERTurk for Turkish.\footnote{\url{https://github.com/stefan-it/turkish-bert}} 
As with Morpheus and IXA pipes, we treat
contextual lemmatization as a sequence tagging task and fine-tune the language
models by adding a single linear layer to the top of the model. The experiments
were implemented using the HuggingFace Transformers
API \citep{Wolf2019TransformersSN}.

\subsection{Systems using morphology}

\noindent IXA pipes is a set of multilingual
tools which is based on a pipeline approach \citep{agerri-etal-2014-ixa,agerri_rigau}. IXA pipes learns 
perceptron \citep{collins-2002} models based on shallow local features combined
with pre-trained clustering features induced over large unannotated corpora. 
The lemmatizer implemented in IXA pipes is inspired by the work of
\citet{chrupala-etal-2008-learning}, where the model learns the SES between the
word form and its lemma. IXA pipes allows to learn lemmatization using
gold-standard or learned morphological tags. 

Morpheus is a neural contextual lemmatizer and morphological tagger which
consists of two separate sequential decoders for generating
morphological tags and lemmas. The input words and morphological features are encoded in context-aware
vector representations using a two-level LSTM network and the decoders predict
both the morphological tags and the SES, which are later decoded into its lemma \citep{yildiz-tantug-2019-morpheus}.
Morpheus obtained the third best overall result in the SIGMORPHON 2019 shared task
\citep{mccarthy-etal-2019-sigmorphon}.

\subsection{Systems without explicit morphological information}

We train a number of models that use modern contextual word representations by
addressing lemmatization as a sequence tagging task. Thus, the input
consists of words encoded as contextual vector representations and the task is
to assign the best sequence of SES to a given input sequence. 

Flair is a NLP framework based on a BiLSTM-CRF architecture
\citep{Huang2015BidirectionalLM,ma-hovy-2016-end} and pre-trained language
models that leverage character-based word representations which, according to
the authors, capture implicit information about natural language syntax and
semantics. Flair has obtained excellent results in sequence labelling tasks
such as named entity recognition, POS tagging and chunking
\citep{akbik-etal-2018-contextual}. The library includes pre-trained Flair language models
for every language except Turkish.

\begin{table}
     \small
    \newcolumntype{Y}{>{\centering\arraybackslash}X}
    \newcolumntype{W}{>{\raggedright\arraybackslash}p{6cm}X}
    \begin{tabularx}{\textwidth}{Y|Y|Y|W} \toprule
        Language &  Model & Architecture & Training corpus and number of tokens \\ \midrule
        Basque & BERTeus & BERT & 35M tokens (Wikipedia) + 191M tokens (online) \\ 
        Czech & slavicBERT & BERT & Russian news and Wikipedia in Russian,\\ & & & Bulgarian, Czech and Polish \\      
        English & RoBERTa & BERT & BookCorpus (800M tokens),\\
        & & & CC-News (16,000M tokens), \\
        & & & OpenWebText (8,706M tokens), \\
        & & & CC-Stories (5,300M tokens)\\ 
        Russian & ruBERT & BERT & Dataset for original BERT (BookCorpus(800M tokens)), \\
        & & &  English Wikipedia (2,500M tokens), \\ 
        & & & Russian news and Wikipedia for subword vocabulary \\ 
        Spanish & BETO & BERT & Wikipedia and OPUS project in Spanish (3,000M tokens)  \\ 
        Turkish & BERTurk & BERT & OSCAR corpus, Wikipedia, OPUS corpora, corpus of Kemal Oflaizer (4,404M tokens total) \\    
        \bottomrule
      \end{tabularx}
    \caption {List of language-specific models used in the experiments for each of the target languages.}
    \label{tab:monolingual_transformers}
    \end{table}

With respect to the MLMs, we use two multilingual models and 6 language models
trained specifically for each of the languages included in our study.
Multilingual BERT \citep{devlin-etal-2019-bert} is a transformer-based masked
language model, pre-trained on the Wikipedias of 104 languages with both the
masking and next sentence prediction objectives. Furthermore, we also use
XLM-RoBERTa \citep{conneau-etal-2020-unsupervised}, trained on 2.5TB (295K millions of tokens) of
filtered CommonCrawl data for 100 languages. XLM-RoBERTa is based on the BERT architecture but (i) trained only on the MLM
task, (ii) on larger batches (iii) on longer sequences and (iv), with dynamic
mask generation. Thus, multilingual BERT was trained with a batch size of 256
and 512 sequence length for 1M steps, using both the MLM and NSP tasks.
Regarding XLM-RoBERTa, both versions (base and large) were trained over 1.5M
steps with batch 8192 and sequences of 512 length.

Details about the six language-specific MLMs used are provided in Table
\ref{tab:monolingual_transformers}. BERTeus \citep{agerri-etal-2020-give} is a
BERT-base model trained on the BMC Basque corpus, which includes the Basque
Wikipedia and news articles from online newspapers. Apart from the training
data, the other difference from original BERT is the subword tokenization,
which is closer to linguistically interpretable strings in Basque. BERTeus
significantly outperforms multilingual BERT and XLM-RoBERTa in tasks such as
POS tagging, named entity recognition, topic modelling and sentiment analysis.

BERTurk\footnote{\url{https://github.com/stefan-it/turkish-bert}} is a cased
BERT-base model for Turkish. This model was trained on a filtered and sentence
segmented version of the Turkish OSCAR corpus
\citep{OrtizSuarezSagotRomary2019}, together with Wikipedia, various OPUS
corpora \citep{tiedemann-2016-opus} and data provided by Kemal Oflazer, which
resulted in total size of 35GB (4,404M tokens total). 

For Czech we used slavicBERT \citep{arkhipov-etal-2019-tuning}, developed
by taking multilingual
BERT as a basis and further pre-trained using Russian news and the Wikipedias
of four Slavic languages: Russian, Bulgarian, Czech and Polish. The authors also
rebuilt the vocabulary of subword tokens, using the subword-nmt repository.\footnote{\url{https://github.com/rsennrich/subword-nmt/}}

RuBERT was developed in a similar fashion as slavicBERT but only with Russian
as target language using the Russian Wikipedia and news corpora \citep{rubert-kuratov}.
They generated a new subword vocabulary obtained from subword-nmt 
which contains longer Russian words and subwords.

For Spanish we used BETO \citep{CaneteCFP2020} -- a BERT-base language model,
trained on a large Spanish corpus. The authors of this model upgraded the
initial BERT model by using the Dynamic Masking technique, introduced in
RoBERTa. BETO performed 2M steps in two different stages: 900K steps with a batch
size of 2048 and maximum sequence length of 128, and the rest of the training
with a batch size of 256 and maximum sequence length of 512.  We use the version
trained with cased data, which included the Spanish Wikipedia and various
sources from the OPUS project \citep{tiedemann-2012-parallel} in a final corpus
size of around 3 billion words. 

RoBERTa-base is the model chosen for English. RoBERTa \citep{liu2019roberta} is an
optimized version of BERT, as commented above. To train this model the authors,
apart from the standard datasets used to train the BERT model, also used the
CC-news dataset, including English news articles from all over the world
published between January 2017 and December 2019. The total size of the
training data exceeds 160GB of uncompressed text (more than 30 billion tokens).

\subsection{Baselines}

We use two models as baselines. First, the system used as a baseline for
the SIGMORPHON 2019 shared task \citep{mccarthy-etal-2019-sigmorphon}, a joint
neural model for morphological tagging and lemmatization presented by
\citet{malaviya-etal-2019-simple}. This system performs morphological tagging by
using a LSTM tagger described in \citet{heigold-etal-2017-extensive} and 
\citet{cotterell-heigold-2017-cross}. The lemmatizer is a neural sequence-to-sequence
model \citep{wu-cotterell-2019-exact} which includes a hard attention mechanism
with a training scheme based on dynamic programming. The tagger and lemmatizer
are connected together by jackknifing \citep{agic-schluter-2017-train}, which
allows to avoid exposure bias and improve lemmatization results.

The second baseline is the winner of the SIGMORPHON'19 shared task
\citep{straka-etal-2019-udpipe}. UDPipe is a multitask model which jointly
learns morphological tagging and lemmatization. The system architecture
consists of three bidirectional LSTMs that process the input and softmax
classifiers that generate lemmas and morphosyntactic features. Lemmatization is
performed as a multiclass classification task, where the system predicts the
correct lemma rule or SES.

\section{Experimental Setup}\label{sec:methodology}

The systems described above were trained on the datasets listed in Section
\ref{sec:datasets} using the following methodology. For the two IXA pipes models (using gold-standard and learned morphology) we
used the default feature set, with and without clustering features, specified
in \citet{agerri_rigau}. The default hyperparameters were also applied to train Morpheus 
\citep{yildiz-tantug-2019-morpheus}. The input character embedding length
$d_{a}$ is set to 128, the length of the word vectors $d_{e}$ to 1024
and the length of the context-aware word vectors $d_{c}$ to 2048. Moreover, 
the length of character vectors in the minimum edit prediction component $d_{u}$
and the length of the morphological tag vectors $d_{v}$ are set to 256. The
hidden unit sizes in the decoder LSTMs $d_{g}$ and $d_{q}$ are set to 1024. The
Adam optimization algorithm is used with learning rate 3e-4 to minimize the
loss \citep{kingma-2015-adam}.

Flair is used off-the-shelf with FastText CommonCrawl word embeddings
\citep{grave-etal-2018-learning} combined with Flair contextual embeddings for
each of the languages. The hidden size of the LSTM is set to 256 with a batch
of 16.

The MLMs were fine-tuned for lemmatization as a sequence tagging task by
adding a single linear layer on top of the model being fine-tuned. A grid search of
hyperparameters was performed to pick the best batch size (16, 32), 
epochs (5, 10, 15, 20, 25) and learning rate (1e-0, 2e-5, 3e-5, 5e-5). We pick the best model on the
development set in terms of word accuracy and loss. A fixed seed is used to ensure reproducibility of the results.

For multilingual BERT we used a maximum sequence length of 128, batch size 32
and 5e-5 as learning rate while for XLM-RoBERTa we used the same configuration but
with a batch of 16. For Russian we perform grid search on two language-specific
models, namely, ruBERT and slavicBERT. RuBERT obtained the best results with a 
maximum sequence length of 128, batch size 16, and a 5e-5 value for learning rate over 15
epochs. For the rest of the models the best configuration was that of
XLM-RoBERTa over 5 epochs for BETO and RoBERTa-base, 10 epochs for BERTeus, 15
epochs for BERTurk and 20 epochs with slavicBERT for Czech.

\section{Experimental Results}\label{sec:results}

In this section we present the experiments to empirically address the following
research questions with respect to the actual role of morphological information to
perform contextual lemmatization, namely, (i) is fine-grained morphological
information really necessary, even for agglutinative languages? (ii) are
modern context-based word representations enough to learn
competitive contextual lemmatizers without including any explicit morphological signal
during training? (iii) do morphologically enriched lemmatizers perform worse
out-of-domain as the complexity of the morphological features increases? (iv)
what is the optimal strategy to obtain robust contextual lemmatizers for
out-of-domain settings? and (v), are current evaluation practices adequate to
meaningfully evaluate and compare contextual lemmatization techniques?

Unlike the vast majority of previous work on contextual lemmatization, which
has been mostly evaluated
in-domain \citep{mccarthy-etal-2019-sigmorphon}, we also report results in
out-of-domain settings. It should be noted that by out-of-domain we
mean to evaluate the model on a different data distribution from the data used
for training \citep{Manning2011PartofSpeechTF}.

First, Section \ref{sec:in-domain} studies the in-domain performance of contextual
lemmatizers depending on the type of morphological features used to inform the
models during training. The objective is two-fold: to determine whether complex
(or any at all) morphological information is required to obtain competitive
lemmatizers and, secondly, to establish whether modern contextual word
representations and MLMs allow us to perform lemmatization without any
morphological information.

Second, in the out-of-domain evaluation presented in Section
\ref{sec:out-of-domain} we analyze the performance of  morphologically informed
lemmatizers. Furthermore, 
comparing them with contextual lemmatizers developed without an explicit morphological
signal would allow us to obtain a full picture as to what is the best strategy
for out-of-domain settings (the most common application scenario).

\subsection{In-domain evaluations}\label{sec:in-domain}

For the first experiment we train the two variants of the IXA pipes statistical
system, ixa-pipe-gs and ixa-pipe-mm
\citep{agerri_rigau}, and one neural lemmatizer, Morpheus
\citep{yildiz-tantug-2019-morpheus}. As explained in Section \ref{sec:systems},
all three require explicit morphological information and they all apply 
the shortest edit script (SES) to automatically induced lemma classes from the
training data.

\begin{table}[hbt!]
    \centering
    \newcolumntype{Y}{>{\centering\arraybackslash}X}
    \begin{tabularx}{\textwidth}{Y}
        \toprule
         Morphological label \\ \midrule
         UPOS \\ 
         UPOS+Case+Gender \\ 
         UPOS+Case+Number \\ 
         UPOS+Case+Gender+Number \\
         UPOS+AllFeaturesOrdered \\ \bottomrule
    \end{tabularx}
    \caption{List of UniMorph morphological tags used.}
    \label{tab:postag-list}
\end{table}

\begin{table}
    \small
     \centering
    \newcolumntype{Y}{>{\raggedright\arraybackslash}X}
    \begin{tabularx}{\textwidth}{Y|Y|Y}
        \toprule
       Word form & Morphological label \{UPOS+Case+\newline{Gender}\} & Lemma \\ \midrule
       Проект\newline{[\emph{Project}]} & NNOMMASC & проект \newline{[\emph{project}]} \\ \midrule
       сильно \newline{[\emph{a lot}]} & ADV & сильно \newline{[\emph{a lot}]}\\ \midrule 
       отличался \newline{[\emph{differed}]}& VMASC & отличаться \newline{[\emph{to differ}]} \\ \midrule
       от \newline{[\emph{from}]}& ADP & от \newline{[\emph{from}]} \\ \midrule
       предыдущих \newline{[\emph{previous}]}& ADJGEN & предыдущий \newline{[\emph{previous}]} \\ \midrule 
       подлодок \newline{[\emph{submarines}]} & NGENFEM & подлодка \newline{[\emph{submarine}]}\\ \midrule
       . & \_\ & . \\ \bottomrule
    \end{tabularx}
    \caption {An example of the data used to train contextual lemmatizers with morphological information.}
    \label{tab:word-label-lemma}
    \end{table}

Furthermore, we combined the UniMorph morphological tags to generate labels of
different complexity. Thus, taking UPOS tags as a basis we obtain 5 different morphological tags, as shown in Table
\ref{tab:postag-list}. The first 4 are combinations of UPOS, case, gender
and number. The last label includes UPOS and every feature present for a given
word in UniMorph in the following order:
\{UPOS+Case+Gender+Number+Rest-of-the-features\}. For some word types, such as
prepositions or infinitives, UniMorph only includes the UPOS tag. In order to
illustrate this, Table
\ref{tab:word-label-lemma} provides an example originally in
Russian including the information required to train contextual lemmatizers,
namely, the word, some morphological tag, and the lemma.

\begin{table}[hbt!]
     \footnotesize
     \centering
    \begin{adjustbox}{min width=\textwidth}
    \begin{tabular}{c|l|r|r|r|r|r|r}\toprule
        & & number & & upos+case & upos & upos &  SES\\  language & corpus &
		of & upos & +gender & +allfeat. & +allfeat. & (lemma) \\ 
        & & tokens & & +number & ord. &
		not.ord. & class)\\ \midrule
        \multirow{4}{3.5em}{\newline Basque} & & & & & & & \\ 
        & train (BDT) & 97,336 & 15 & 205 & 1,143 & 1,683 & 1,306 \\
        & dev (BDT) & 12,206 & 14 & 148 & 556 & 787 & 432 \\
        & test (BDT) & 11,901 & 14 & 153 & 545 & 773 & 428\\
        & test (Armiarma) & 299,206 & - & - & - & - & 1,495\\ \midrule
        \multirow{4}{3.5em}{\newline Czech} & & & & & & & \\ 
        & train (CAC) & 395,043 & 16 & 332 & 1,266 & 1,784 & 946\\ 
        & dev (CAC) & 50,087 & 16 & 298 & 876 & 1,129 & 536\\
        & test (CAC) & 49,253 & 15 & 284 & 827 & 1,036 & 556\\
        & test (PUD) & 1,930 & 14 & 175 & 288 & 292 & 151\\ \midrule
        \multirow{5}{3.5em}{\newline Russian} & & & & & & & \\ 
        & train (GSD) & 79,989 & 14 & 241 & 851 & 1,384 & 553\\ 
        & dev (GSD) & 9,526 & 14 & 191 & 435 & 673 & 235\\
        & test (GSD) & 9,874 & 14 & 203 & 455 & 713 & 258\\
        & test (SynTagRus) & 109,855 & 15 & 247 & 757 & 1,243 & 896\\ \midrule
        \multirow{4}{3.5em}{\newline Spanish} & & & & & & & \\ 
        & train (GSD) & 345,545 & 25 & 116 & 287 & 510 & 310\\ 
        & dev (GSD) & 42,545 & 23 & 100 & 208 & 342 & 200\\
        & test (GSD) & 43,497 & 23 & 103 & 222 & 387 & 200\\
        & test (AnCora) & 54,449 & 15 & 75 & 178 & 309 & 298\\ \midrule
         \multirow{5}{3.5em}{\newline English} & & & & & & & \\ 
        & train (EWT) & 204,857 & 16 & 43 & 94 & 173 & 233\\ 
        & dev (EWT) & 24,470 & 16 & 41 & 88 & 160 & 120\\
        & test (EWT) & 25,527 & 16 & 41 & 85 & 156 & 115\\
        & test (GUM) & 8,189 & 17 & 42 & 72 & 124 & 80\\ \midrule
         \multirow{2}{3.5em}{\newline \newline Turkish} & & & & & & & \\ 
        & train (IMST) & 46,417 & 15 & 124 & 1,541 & 1,897 & 211 \\ 
        & dev (IMST) & 5,708 & 15 & 95 & 605 & 748 & 106\\ 
        & test (IMST) & 5,734 & 16 & 100 & 589 & 725 & 104\\
        & test (PUD) & 1,795 & 15 & 66 & 217 & 220 & 59\\ \bottomrule
    \end{tabular}
    \end{adjustbox}
     \caption {Language complexity reflected in the number of labels according to the augmentation of morphological features, number of lemma classes and corpus tokens.}
    \label{tab:abanico-de-idiomas}
    \end{table}

Putting it all together, Table \ref{tab:abanico-de-idiomas} characterizes the
final datasets used for in- and out-of-domain evaluation. The number of tokens,
unique labels per category and unique SES (calculated using the UDPipe method)
illustrate the varied complexity of the languages involved.\footnote{Even though
it is not required for out-of-domain evaluation, the UniMorph information
is not available for the Basque Armiarma corpus because it is not part of the
UniMorph project.} Thus, those languages with more complex morphology have a higher number of unique
labels that include additional morphological features. The same pattern can be
seen in the amount of lemma classes
(SES), significantly larger for the languages with more complex morphology. In
the case of Turkish the low number of lemmas could be explained by the fact
that most Turkish words are formed by applying derivative suffixes to nouns
and verbal stems. Moreover, the core vocabulary in this particular corpus is
rather small. Finally, we decided to order the subtags comprising the full 
UniMorph labels as the number of unique labels decreased significantly. 

Table \ref{tab:fine-grained-morphology-development} reports the in-domain
results of training the three systems for the six languages with the 5
different types of morphological labels. First, the results show that the
neural lemmatizer Morpheus outperforms the statistical lemmatizers for every
language except English. In fact, for languages with more complex morphology,
such as Basque and Turkish, the differences are larger. Second, if we look at
the impact of including fine-grained morphological features it can be seen that
no single morphological tag performs best across systems and languages. Thus,
while adding case, number and/or gender seems to be slightly beneficial,
differences in performance are substantial when training the statistical
lemmatizer using gold-standard morphological labels (ixa-pipe-gs) and
especially for languages with more complex morphology (Basque, Russian,
Turkish). Third, the results clearly show that adding every available
morphological feature is not beneficial per se. Fourth, the statistical
lemmatizer trained with learned morphological tags (ixa-pipe-mm) performs
significantly worse in every case except for English and Spanish. Finally,
adding a special label `no-tag' with no morphological information shows that
performance decreases significantly for every system and language.

\begin{table}[h]
	\small
     \centering
    \newcolumntype{Y}{>{\centering\arraybackslash}X}
    \begin{tabularx}{\textwidth}{Y|YYYYYY}\toprule
    \multicolumn{7}{c}{English} \\ \midrule
           & no-tag & UPOS & UCG & UCN & UCGN & UAllo \\ \midrule
    ixa-mm  & - & 98.97 & 98.97 & 99.03 & 98.97 & 98.86 \\
    ixa-gs & 96.98 & 99.51 & 99.49 & 99.58 & 99.59 & \underline{99.65} \\
    morph & 97.60 & 98.20 & 98.12 & 98.13 & 98.19 & 98.14 \\ \midrule
    \multicolumn{7}{c}{Spanish} \\ \midrule
    ixa-mm  & - & 98.75 & 98.74 & 98.71 & 98.78 & 98.74 \\
    ixa-gs & 98.36 & 98.82 & 98.78 & 98.82 & 98.80 & 98.88 \\
    morph & 98.17 & 98.09 & 98.93 & \underline{98.96} & 98.92 & 98.91 \\
     \midrule
    \multicolumn{7}{c}{Russian} \\ \midrule
    ixa-mm  & - & 94.85 & 95.37 & 95.69 & 95.50 & 95.53 \\
    ixa-gs & 91.85 & 95.05 & 96.95 & 96.45 & 96.99 & 97.04 \\
    morph & 96.50 & 96.92 & 96.91 & 97.10 & 97.18 & \underline{97.24} \\
     \midrule
    \multicolumn{7}{c}{Basque} \\ \midrule
    ixa-mm  & - & 93.19 & 93.22 & 93.14 & 93.30 & 93.49 \\
    ixa-gs & 91.68 & 93.50 & 94.33 & 94.58 & 94.58 & 96.50 \\
    morph & 95.48 & 96.30 & 96.43 & \underline{96.54} & 96.37 & 96.42 \\
     \midrule
    \multicolumn{7}{c}{Czech} \\ \midrule
    ixa-mm  & - & 97.76 & 97.17 & 97.29 & 97.10 & 97.10 \\
    ixa-gs & 95.64 & 97.68 & 98.10 & 97.93 & 98.09 & 98.20 \\
    morph & 98.37 & 98.78 & \underline{98.84} & 98.83 & 98.82 & 98.80 \\
     \midrule
    \multicolumn{7}{c}{Turkish} \\ \midrule
    ixa-mm  & - & 84.83 & 84.51 & 85.06 & 85.06 & 83.95 \\
    ixa-gs & 85.97 & 88.81 & 88.89 & 89.14 & 89.14 & 90.52 \\
    morph & 96.04 & 96.41 & \underline{96.53} & 95.95 & 96.27 & 96.50 \\
     \bottomrule
     \end{tabularx}
     \caption{In-domain lemmatization results on the development sets for systems that use morphology to train contextual lemmatizers. ixa-mm: IXA pipes with learned morphological tags; ixa-gs: IXA pipes with gold standard morphology; morph = Morpheus; UCG: UPOS+Case+Gender, UCN: UPOS+Case+Number, UCGN:UPOS+Case+Gender+Number: UALLo: UPOSAllFeaturesOrdered.}
    \label{tab:fine-grained-morphology-development}
    \end{table}    

Summarizing, in-domain performance for high-inflected languages improves
when some fine-grained morphological attributes (case and number or gender) are used to
train the statistical lemmatizers. However, for English and Spanish using UPOS
seems to be enough. Thus, in the case of neural lemmatization with Morpheus
(the best of the models using morphological information), we can see that no substantial gains
are obtained by adding fine-grained morphological features to UPOS tags, not
even for agglutinative languages such as Basque or Turkish. 

This point is reinforced by the results of computing the McNemar test of
statistical significance to establish whether the differences in the
results obtained by Morpheus (the best among the models trained with morphology)
informed only with UPOS labels or with the best morphological label (as by
Table \ref{tab:fine-grained-morphology-development} above) are statistically
significant or not (null hypothesis). The result of the test showed that for
every language the differences were not significant ($\alpha = .05$, with
0.936 p-value for Basque, 0.837 for Czech, 0.511 for Russian and 0.942 for
Spanish). 



Taking this into consideration, the next natural step is to consider
whether it is possible to learn good contextual lemmatizers without providing
any explicit morphological signal during training. Previous work on probing
contextual word representations and transformer-based masked language models
(MLMs) suggests that such
models implicitly encode information about part-of-speech and morphological
features
\citep{Manning2020EmergentLS,akbik-etal-2018-contextual,conneau-etal-2018-cram,Belinkov2017WhatDN}.
Following this, for this experiment we fine-tune various well-known multilingual and monolingual
language models (detailed in Section \ref{sec:systems}) by using only the word
forms and the automatically induced shortest edit scripts (SES) as implemented by
UDPipe \citep{straka-etal-2019-udpipe}.

\begin{figure}
	\centering
        \includegraphics[width=39em]{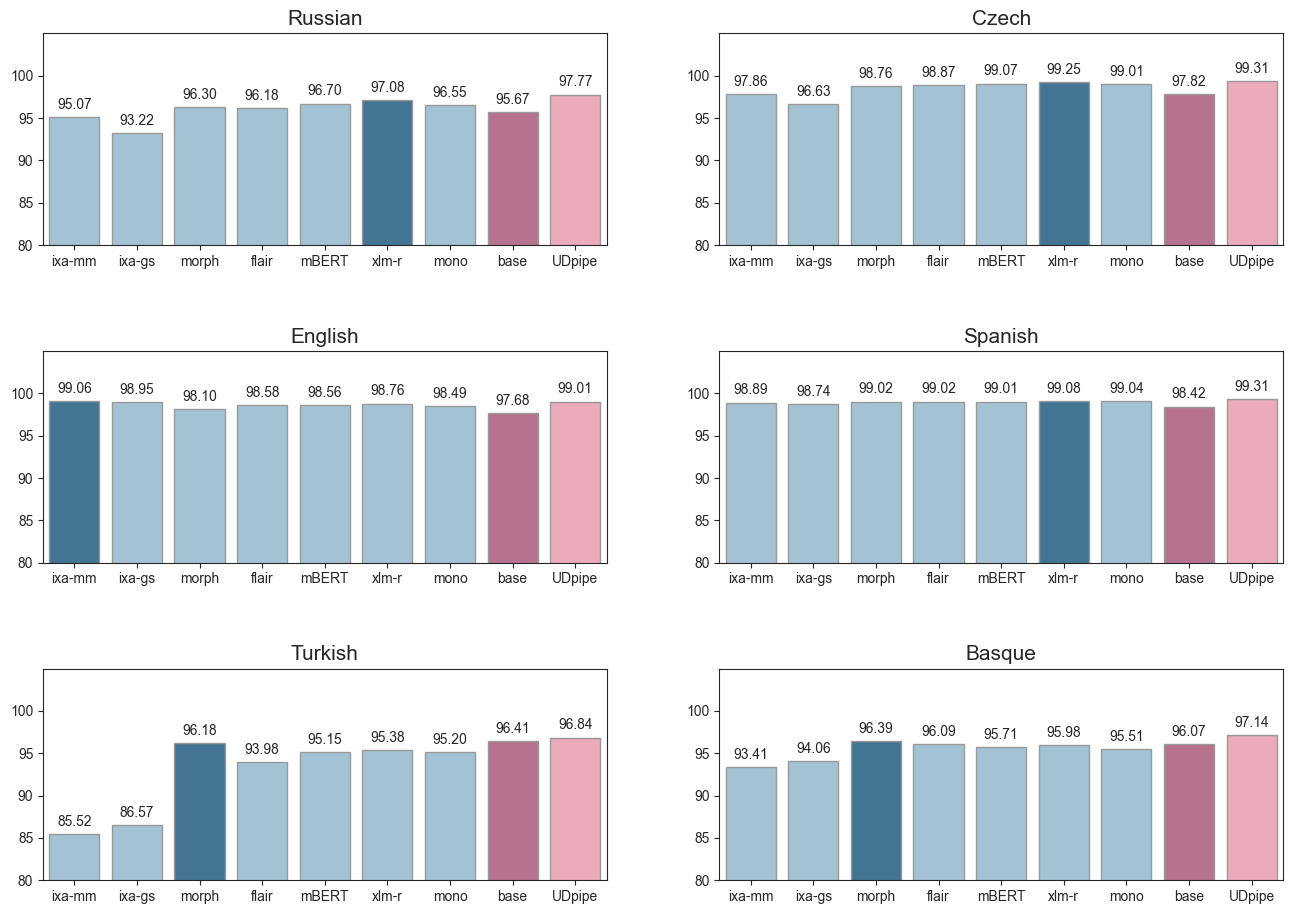}
        \caption{Overall in-domain
		lemmatization results on the test data for models trained with and without explicit
	morphological features; monolingual transformers: Russian - ruBERT, Czech - slavicBERT, Basque - BERTeus, Turkish - BERTurk, English - RoBERTa, Spanish - BETO.}
\label{fig:in_domain_overall.png} 
\end{figure}

Figure \ref{fig:in_domain_overall.png} reports the results. From left-to-right,
the first three bars correspond to the best statistical and Morpheus models
using explicit morphological information as previously reported in Table
\ref{tab:fine-grained-morphology-development}. The next four list the
results from Flair, mBERT, XLM-RoBERTa-base and a language-specific monolingual
model (none of these four use any explicit morphological signal) whereas base (dark purple) refers to
the system of \citet{malaviya-etal-2019-simple}, employed as a baseline for the SIGMORPHON
2019 shared task \citep{mccarthy-etal-2019-sigmorphon}. 
For state-of-the-art comparison, the last column on the right provides the results from UDPipe
\citep{straka-etal-2019-udpipe} (light purple color). Finally, the dark blue bars represent the best result for each language without
considering either the baseline system or UDPipe.

The first noticeable trend is that every model beats the baseline except the
IXA pipes-based statistical lemmatizers, which perform over the baseline and
comparatively to the other models for English and Spanish only, the languages
with the less complex morphology. 

The second and, perhaps, most important fact is that the four models (Flair, mBERT, XLM-RoBERTa and mono) which do not use
any morphological signal for training, obtain a remarkable performance across
languages, XLM-RoBERTa-base being the best overall, even better than
language-specific monolingual models. In fact, XLM-RoBERTa-base outperforms Morpheus for 4
out of the 6 languages, a neural model which was the third best system in the
SIGMORPHON 2019 benchmark and which uses all the morphological information
available in the UniMorph data. The McNemar test of significance shows that the
differences in results obtained by Morpheus and XLM-RoBERTa are statistically
significant ($\alpha = .05$) for Russian, Spanish and English (in
XLM-RoBERTa's favour), and for Basque and Turkish (Morpheus over XLM-RoBERTa).

An additional observation is our XLM-RoBERTa-base lemmatization models perform
competitively with respect to UDPipe, which obtains the best results for 5 out
of the 6 languages included in our study. UDPipe's strong performance is
somewhat expected as it was the overall winner of the SIGMORPHON 2019
lemmatization task. It should be noted that UDPipe is a rather complex system
consisting of a multitask model to predict POS tags, lemmas and dependencies by
applying three shared bidirectional LSTM layers which take as input a variety
of word and character embeddings, the final model being an ensemble of 9
possible embedding combinations. However, the results obtained by the language
models we trained without any explicit morphological signal, such as
XLM-RoBERTa-base, are based on a simple baseline setting, where the transformer
models are fine-tuned using the automatically induced SES as the target labels
in a token classification task. These results seem to confirm that, as it was
the case for POS tagging and other tasks \citep{Manning2020EmergentLS},
contextual word representations implicitly encode morphological information
which made them perform strongly for lemmatization.

\begin{table}
    \scriptsize	
    \centering
    \newcolumntype{Y}{>{\centering\arraybackslash}X}
    \begin{tabularx}{\textwidth}{Y|YYY|YYY}\toprule
    & \multicolumn{3}{c}{IN-DOMAIN} & \multicolumn{3}{c}{OUT-OF-DOMAIN} \\ \midrule
    \multirow{3}{*}{} &
    \multicolumn{6}{c}{NO TAG} \\ \midrule
    & ixa-mm & ixa-gs & morph & ixa-mm & ixa-gs & morph \\
	en & - & 96.34 & \underline{97.51} & - & 90.40 & \underline{92.47}\\
	es & - & \underline{98.53} & 98.17 & - & \underline{89.75} & 89.70\\
	ru & - & 92.81 & \underline{95.31} & - & 83.95 & \underline{86.84}\\
	eu & - & 90.61 & \underline{95.69} & - & 85.64 & \underline{88.25}\\
	cs & - & 96.37 & \underline{98.31} & - & 91.50 & \underline{91.61}\\
	tr & - & 87.11 & \underline{95.62} & - & 77.16 & \underline{84.07}\\ \midrule
    & \multicolumn{6}{c}{UPOS} \\ \midrule
    & ixa-mm & ixa-gs & morph & ixa-mm & ixa-gs & morph \\
	en & \underline{99.11}$^*$ & 98.91 & 98.10 & \underline{95.38}$^*$ & 95.25 & 92.92\\
	es & 98.91 & 98.76 & \underline{98.94} & \underline{97.53} & 97.41 & 90.29\\
	ru & 94.36 & 93.74 & \underline{96.20} & \underline{90.00} & 89.40 & 87.59 \\
	eu & 93.11 & 92.29 & \underline{96.39} & 85.22 & 86.79 & \underline{88.97}\\
	cs & 97.86 & 97.28 & \underline{98.75} & 92.33 & \underline{93.68} & 91.66 \\
	tr & 84.65 & 87.76 & \underline{96.44}$^*$ & 79.22 & 81.67 & \underline{84.96}$^*$ \\ \midrule
    & \multicolumn{6}{c}{UCG}  \\ \midrule
     & ixa-mm & ixa gs & morph & ixa-mm & ixa-gs & morph \\
	en & \underline{99.10} & 98.92 & 97.99 & 95.20 & \underline{95.24} & 92.97\\
	es & 98.94 & 98.70 & \underline{98.98} & \underline{97.54} & 97.43 & 90.31\\
	ru & 94.85 & 93.30 & \underline{96.21} & 90.97 & 89.33 & 87.67 \\
	eu & 92.65 & 92.39 & \underline{96.34} & 85.23 & 86.74 & \underline{89.09}\\
	cs & 97.29 & 96.64 & \underline{98.76}$^*$ & 91.61 & 91.35 & \underline{91.92} \\
     tr & 85.09 & 87.09 & 96.18 & 80.06 & 81.23 & 84.74 \\ \midrule
    & \multicolumn{6}{c}{UCN} \\ \midrule
    & ixa-mm & ixa-gs & morph & ixa-mm & ixa-gs & morph \\
	 en & \underline{99.06} & 98.87 & 98.01 & \underline{95.16} & 95.16 & 92.86 \\
	 es & 98.92 & 98.75 & \underline{99.02}$^*$ & \underline{97.56} & 97.44 & 90.35\\
	 ru & 95.07 & 93.70 & \underline{96.20} & \underline{91.00}$^*$ & 89.60 & 87.58 \\
	 eu & 93.03 & 92.35 & \underline{96.39} & 85.47 & 86.36 & \underline{89.03} \\
	 cs & 97.44 & 96.87 & \underline{98.71} & 91.04 & 92.07 & \underline{92.23}$^*$ \\
	 tr & 85.52 & 87.18 & \underline{96.11} & 80.33 & 81.00 & \underline{84.40} \\ \midrule
    & \multicolumn{6}{c}{UCGN} \\ \midrule
     & ixa-mm & ixa-gs & morph & ixa-mm & ixa gs & morph \\
	 en & \underline{99.08} & 98.96 & 97.99 & \underline{95.21} & 95.15 & 92.95\\
	 es & 98.89 & 98.71 & \underline{98.97} & \underline{97.59}$^*$ & 97.44 & 90.38 \\
	 ru & 95.00 & 93.08 & \underline{96.44}$^*$ & \underline{90.80} & 89.13 & 87.66 \\
	 eu & 93.03 & 92.28 & \underline{96.39} & 85.38 & 86.55 & \underline{88.86} \\
	 cs & 97.17 & 96.68 & \underline{98.70} & 91.71 & 91.50 & \underline{91.97} \\
	 tr & 85.52 & 87.18 & \underline{96.20} & 80.33 & 81.00 & \underline{84.46}\\ \midrule
    & \multicolumn{6}{c}{UAllo} \\ \midrule
     & ixa-mm & ixa-gs & morph & ixa-mm & ixa-gs & morph \\
	 en & \underline{99.04} & 98.95 & 98.06 & 95.08 & \underline{95.13} & 93.15 \\
	 es & 98.86 & 98.74 & \underline{99.00} & \underline{97.54} & 97.45 & 90.34 \\
	 ru & 94.75 & 93.22 & \underline{96.30} & \underline{90.88} & 88.66 & 87.57 \\
	 eu & 93.41 & 94.06 & \underline{96.50}$^*$ & 85.33 & 86.31 & \underline{89.11}$^*$\\
	 cs & 97.03 & 96.63 & \underline{98.70} & 91.19 & 91.81 & \underline{92.02}\\
	 tr & 84.90 & 86.57 & \underline{96.22} & 79.39 & 80.50 & \underline{84.96} \\
     \bottomrule
    \end{tabularx}
    \caption{In-domain and out-of-domain test results for systems trained with explicit morphological information: ixa-mm: IXA pipes with learned morphological tags; ixa-gs: IXA pipes with gold standard morphology; morph: Morpheus; UCG: UPOS+Case+Gender, UCN: UPOS+Case+Number, UCGN: UPOS+Case+Gender+Number: UALLo: UPOSAllFeaturesOrdered. \underline{Underline}: Best model per language and type of label; $^*$: best overall per language.}
    \label{tab:in-out-detailed_morph_results}
    \end{table}    

However, we can see that for agglutinative languages such as Basque and
Turkish, the neural models using explicit morphological features (Morpheus,
Malaviya et al. 2019 and UDPipe) still outperform those without it (although
for Basque the differences are much smaller). Still, the overall results show
that, apart from Basque and Turkish, differences between XLM-RoBERTa and
the best model for each language are rather minimal.  This demonstrates that it
is possible to generate competitive contextual lemmatization without any
explicit morphological information using a very simple technique, although a
more sophisticated approach or larger language model may be required to be
competitive with the state-of-the-art currently represented by UDPipe. 

\subsection{Out-of-domain evaluation}\label{sec:out-of-domain}


Although lemmatizers are mostly used out-of-domain, the large majority of the
experimental results published so far do not take this issue into account when
evaluating approaches to contextual lemmatization. In this section we
empirically investigate the out-of-domain performance of the lemmatizers from
the previous section to establish whether: (i) using fine-grained morphological
information causes cascading errors in the lemmatization performance; (ii)
whether the lack of morphological information helps to obtain more robust
lemmatizers across domains. 

For a better comparison, Table \ref{tab:in-out-detailed_morph_results} presents
both the in-domain results presented in the previous section together with
their corresponding out-of-domain performance on the datasets presented in
Section \ref{sec:languages_and_datasets}.

Table \ref{tab:in-out-detailed_morph_results} allows to see the general trend
in performance across domains and with respect to the type of morphological
information used. First, and as it could be expected, out-of-domain performance
is substantially worse for every evaluation setting and particularly
significant for highly-inflected languages. Second, in terms of the type of
morphological label, there are no clear differences between the models using
just UPOS tags or those using more fine-grained information, the exception
being Russian and Turkish with the ixa-pipe-mm system, for which the
highest result with \{UPOS+Case+Number\} is around 1 point in word accuracy
better than UPOS. Furthermore, there is not a common type of morphological
information that works best across languages. Third, while the statistical
lemmatizers are competitive for Spanish and English, they are clearly inferior
for Basque and Turkish. Finally, when looking at the results in terms of the
models using gold-standard morphological annotations (ixa-pipe-gs and
Morpheus) it is interesting that they degrade less out-of-domain than the model
using learned morphological tags for most of the cases except for Russian.
Summarizing, we can conclude that adding fine-grained morphological information
to UPOS does not in general result in better out-of-domain performance.

\begin{figure}
	\centering
	\includegraphics[width=39em]{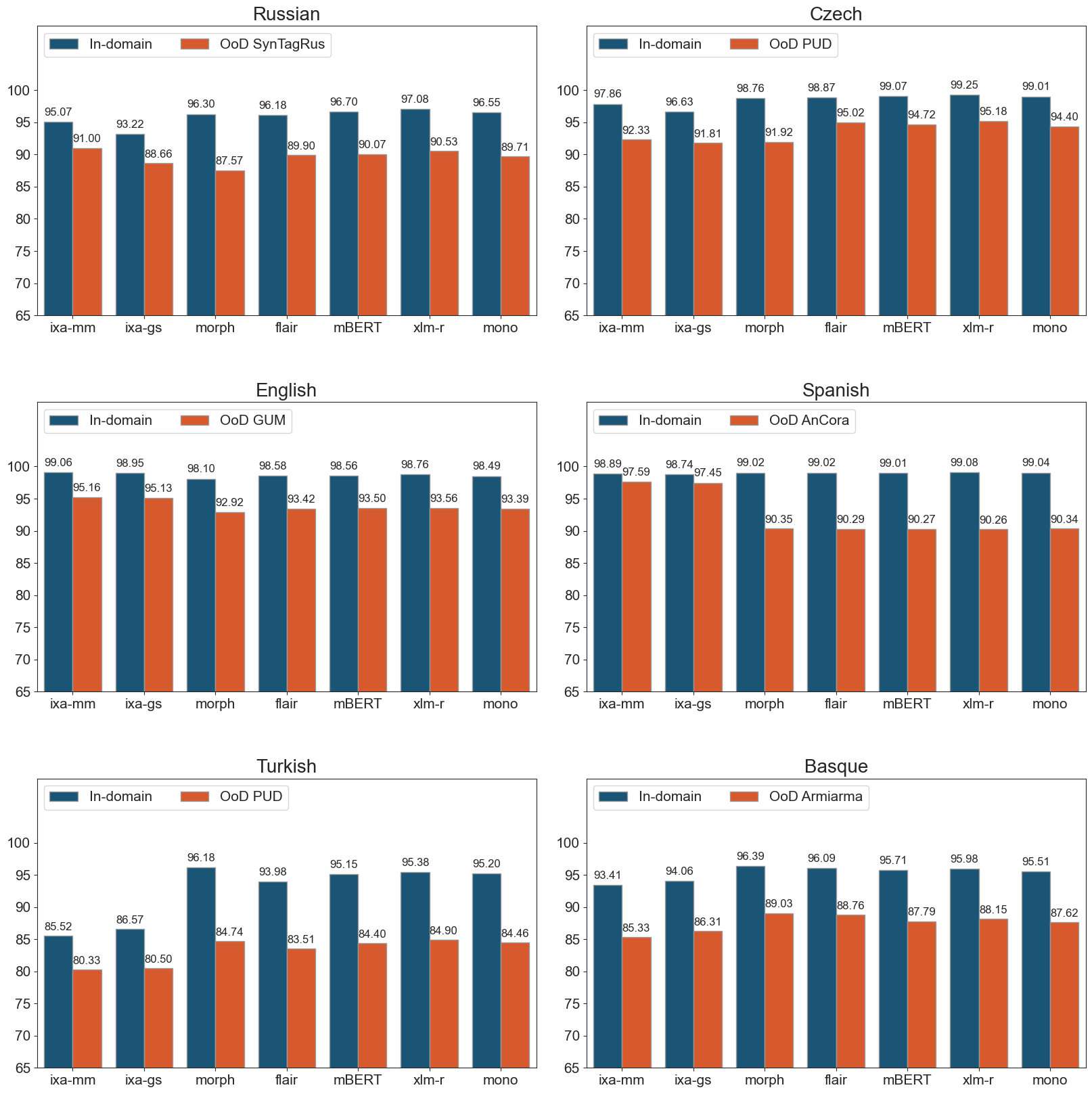}
	\caption{Overall in-domain and out-of-domain results.}
	\label{fig:inD_outD_overall.png} 
\end{figure}

\begin{figure}
	\centering
	\includegraphics[width=39em]{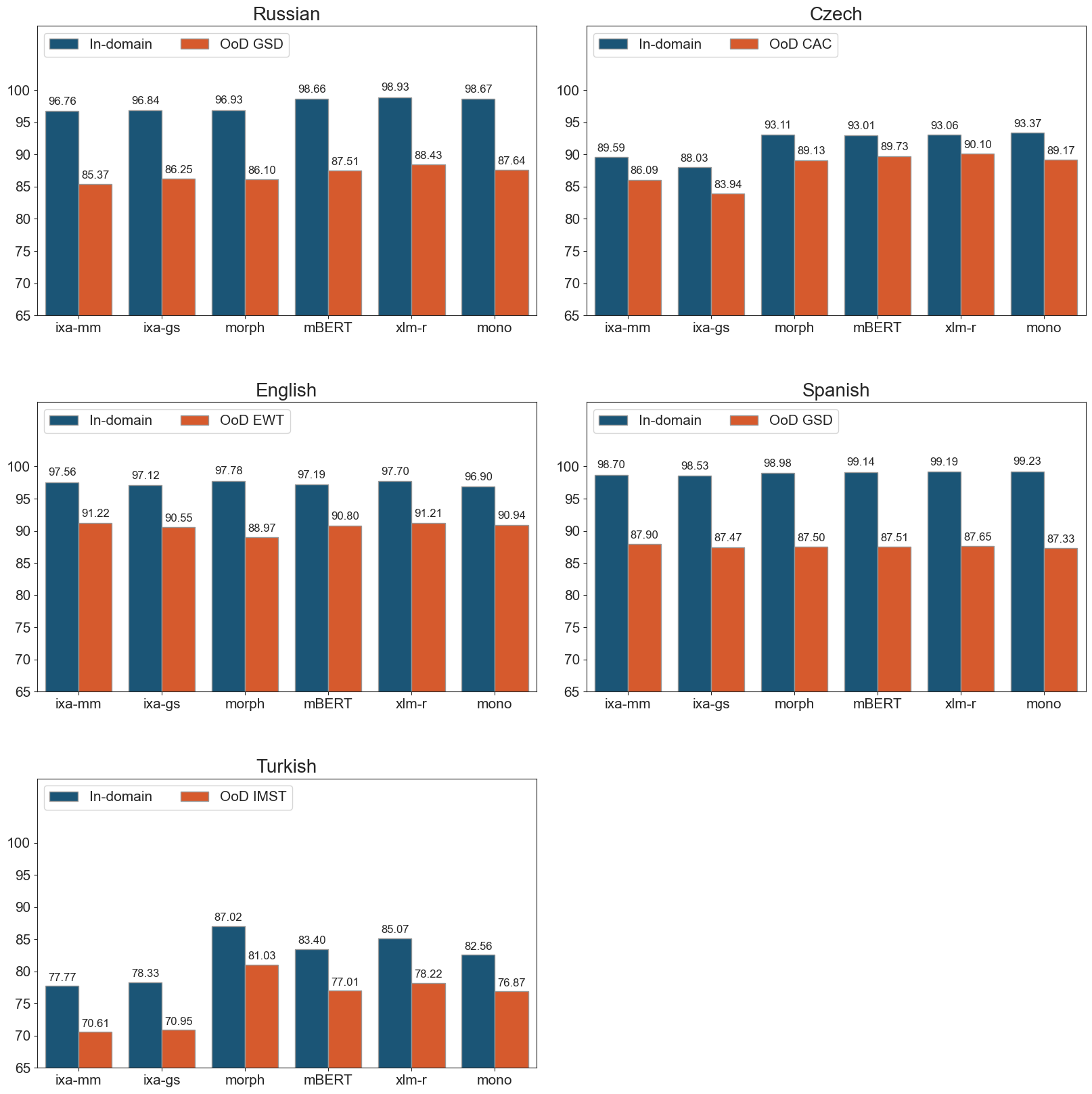}
	\caption{Overall in- and out-of-domain results in the reversed setting.}
\label{fig:in_domain_out_of_domain_overall_reversed.png}
\end{figure}

Following this, we would like to evaluate the out-of-domain performance when
not even UPOS labels are used for training. From what we have seen in-domain,
the systems that operate without morphology achieve competitive results with
respect to the models using morphological information. Figure
\ref{fig:inD_outD_overall.png} provides an overview of both the in- and
out-of-domain results obtained for
both types of systems, confirming this trend. Thus, it is remarkable that the
XLM-RoBERTa model scores best out-of-domain for Turkish and Czech, and a very
close second in Russian. The results for Spanish and English deserve further
analysis, as the IXA pipes statistical models clearly outperform every other
system for these two languages, with the differences around 7 points in word
accuracy. 

Figure \ref{fig:in_domain_out_of_domain_overall_reversed.png}\footnote{Basque
is not present in this evaluation due to the fact that the Armiarma corpus does
not include UniMorph annotations.} presents the
reversed results of those presented in Figure \ref{fig:inD_outD_overall.png},
namely, the test set of the in-domain corpora becomes the out-of-domain test
data while the models are fine-tuned on the training split of the out-of-domain
data. Doing this experiment allows to discard that the out-of-domain behaviour
exhibited in previous results could be due to differences in size between the
training in-domain data and the testing out-of-domain test sets. Good examples
of this are Russian and Spanish for which SynTagRus and AnCora
are used as in-domain data in the reversed setting. These two datasets are much larger
than the GSD corpora for those languages (used as in-domain data in the
original setting). Thus, results in the reversed setting demonstrate that: (i) out-of-domain performance worsens
substantially regardless of the language and model, (ii) language models
fine-tuned without explicit morphological information outperform in-domain every other
model for all languages except Turkish, and (iii), the out-of-domain results of
XLM-RoBERTa-base are the best for Russian and Czech and similar to other models in
English and Spanish.

In any case, Figures \ref{fig:inD_outD_overall.png} and
\ref{fig:in_domain_out_of_domain_overall_reversed.png} show that the results of every model
significantly degrade when evaluated out-of-domain, the most common
application of lemmatizers. Thus, even for high-scoring languages such as
English and Spanish, out-of-domain performance worsens between 3 and 5 points
in word accuracy. For high-inflected languages the differences are around 8 for
Basque and more than 10 for Turkish.

Given that pre-trained language models such as XLM-RoBERTa-base can be leveraged to
learned competitive lemmatizers without using any explicit morphological
signal, we propose a final experiment to address the following two additional
research questions. First, will lemmatization results get closer to the
state-of-the-art by using a larger transformer-based model such as
XLM-RoBERTa-large? Second, can we improve the performance of a language model
such as XLM-RoBERTa by adding morphological information during fine-tuning?

\begin{table}[h]
\small
\newcolumntype{Y}{>{\centering\arraybackslash}X}
\begin{tabularx}{\textwidth}{Y|YY|YY|YY|YY}
	\toprule
          & \multicolumn{4}{c}{xlm-r base} & \multicolumn{4}{c}{xlm-r large}\\ \midrule
          & \multicolumn{2}{c}{in-domain}  & \multicolumn{2}{c}{out-of-domain}
		  & \multicolumn{2}{c}{in-domain} & \multicolumn{2}{c}{out-of-domain} \\ \midrule
          & without & with & without & with & without & with & without & with \\
          & morph. & morph. & morph. & morph. & morph. & morph. & morph. & morph. \\ \midrule
	en  & 98.76  & 98.74 & 93.56  & 93.72 & 98.85  & \underline{98.92} & 93.82 & \underline{93.86}  \\
	es & 99.08 & 99.10 & 90.26 & 90.42 & 99.12 & \underline{99.15} & 90.48 & \underline{90.53}  \\
	eu & 95.98 & 96.45 & 88.15  & 88.60  & 96.66 & \underline{96.70} & 88.75 & \underline{88.81}  \\
	ru & 97.08  & 97.25 & 90.53 & 90.92 & 97.63 & \underline{97.96} & 91.60 & \underline{91.71}   \\
	cz  & 99.25 & 99.32 & 95.18 & 94.72 & \underline{99.40} & 99.23 & 95.42 & \underline{96.06}  \\
	tr & 95.38 & 95.19 & 84.90 & 85.34 & \underline{96.30} & 96.13 & 85.18 & \underline{85.40}   \\ \bottomrule
\end{tabularx}
\caption{In- and out-of-domain results for XLM-RoBERTa-base and
XLM-RoBERTa-large models with and without morphological features during
training.}
\label{tab:xlmr-morphology}
\end{table}

Table \ref{tab:xlmr-morphology} shows the results of experimenting with
XLM-RoBERTa-base and XLM-RoBERTa-large to learn lemmatization as a sequence
labelling task with and without adding morphology as explicit handcrafted
features. For each language we pick the best morphological configuration from
Table \ref{tab:in-out-detailed_morph_results} and encode the morphological labels as feature embeddings.
Both feature and encoded text embeddings are then sent into a softmax layer for sequence labelling \citep{yang2022yato}.
The first observation is that the large version of XLM-RoBERTa obtains the best results both in- and out-of
domain. It is particularly noteworthy that fine-tuning XLM-RoBERTa-large with
only the SES classes helps to outperform any other model for every language
and evaluation setting. Furthermore, adding morphology as a
feature seems to be beneficial. In fact, the morphologically
informed models are the best in 4 out of 6 in-domain evaluations and for all 6
out-of-domain cases.  

We compute the McNemar test to establish whether the differences obtained
with and without morphological features are actually statistically significant.
It turns out that for XLM-RoBERTa-large results are rather mixed. Thus, 
only for Russian (p-value 0.003) and Czech (0.000) are the results significant at $\alpha = .05$. For Turkish and Basque the results are not conclusive (p-value 0.0495) while
for the rest the null hypothesis cannot be rejected (0.423 for Spanish, 0.242
in English and 0.547 in Basque). Regarding XLM-RoBERTa-base, in 4 out of 6 languages
the results are statistically significant at $\alpha = .01$ (the McNemar test),
failing to reject the null hypothesis for Russian and Turkish.

To sum up, our experiments empirically demonstrate that fine-grained
morphological information to train contextual lemmatizers does not lead to
substantially better in- or out-of-domain performance, not even for languages of varied
complex morphology, such as Basque, Czech, Russian and Turkish. Thus, only for
Basque and Turkish did Morpheus (using UPOS tags) outperformed XLM-RoBERTa
models.

Taking this into account, and as previously hypothesized for other NLP tasks \citep{Manning2020EmergentLS}, modern
contextual word representations seem to implicitly capture morphological
information valuable to train lemmatizers without requiring any explicit
morphological signal. We have proved this by training off-the-shelf language models to
perform lemmatization as a token classification task obtaining state-of-the-art
results for Russian and Czech, and very close performance to UDPipe in the
rest. Finally, statistical models are only competitive to perform contextual lemmatization on languages
with a morphology on the simple side of the complexity spectrum, such as
English or Spanish.

Thus, the results indicate that XLM-RoBERTa-large is the optimal option to
learn lemmatization without any explicit morphological signal for every
language and evaluation setting.

\section{Discussion}\label{sec:discussion}

In this paper we performed a number of experiments to better understand the
role of morphological information to learn contextual lemmatization. Our findings can be summarized as follows: (i)
fine-grained morphological information does not help to substantially improve contextual
lemmatization, not even for high-inflected languages; using UPOS tags seems to
be enough for comparable performance; (ii) contextual word representations such as those
employed in transformer and Flair models seem to encode enough implicit
morphological information to allow us to train good performing lemmatizers without any explicit
morphological signal; (iii) the best-performing lemmatizers out-of-domain are
those using either simple UPOS tags or no morphology at all; (iv) evaluating
lemmatization on word accuracy is not the best strategy; results are too high
and too similar to each other to be able to discriminate between models.
By using word accuracy we are assigning the same importance to cases in which
the lemma is equivalent to the word form (e.g. `the') as to complex cases in which the word
form includes case, number and/or gender information (e.g, `medikuarenera',
which in Basque means ``to the doctor'', with its corresponding lemma `mediku'). We believe that this may lead to a
high overestimation in the evaluation of the lemmatizers.

In this section, we address some remaining open
issues  with the aim of understanding better the main errors and difficulties still facing
lemmatization. First, we discuss the convenience of using an alternative metric to word accuracy. 
Second, we analyze the performance of XLM-RoBERTa-base by evaluating
accuracy per SES. Third, we examine the generalization capabilities of
XLM-RoBERTa-base by computing word accuracy for in-vocabulary and
out-of-vocabulary words. We also discuss any issues regarding test data
contamination. Finally, we perform some error analysis on the out-of-domain
performance of the XLM-RoBERTa-base model for Spanish, to see why it is different to
the rest of the languages, as illustrated by Figure
\ref{fig:inD_outD_overall.png}.

\subsection{Sentence Accuracy}\label{sec:sentence_accuracy}

Looking at the in-domain results for lemmatization reported in the previous
sections and in the majority of recent work
\citep{malaviya-etal-2019-simple,mccarthy-etal-2019-sigmorphon,yildiz-tantug-2019-morpheus,straka-etal-2019-udpipe},
with word accuracy in-domain scores around 96 or higher, it is not surprising
to wonder whether contextual lemmatization is a solved task. However, if we
look at the evaluation method a bit more closely, things are not as clear as they seem. As
it has been argued for POS tagging \citep{Manning2011PartofSpeechTF}, word
accuracy as an evaluation measure is easy because you get many free points for
punctuation marks and for the many tokens that are not ambiguous with respect to
its lemma, namely, those cases in which the lemma and the word form are the
same. Following this, a more realistic metric might consist of looking at the
rate of getting the whole sentence correctly lemmatized, just as it was
proposed for POS tagging \citep{Manning2011PartofSpeechTF}. 

\begin{figure}
	\includegraphics[width=39em]{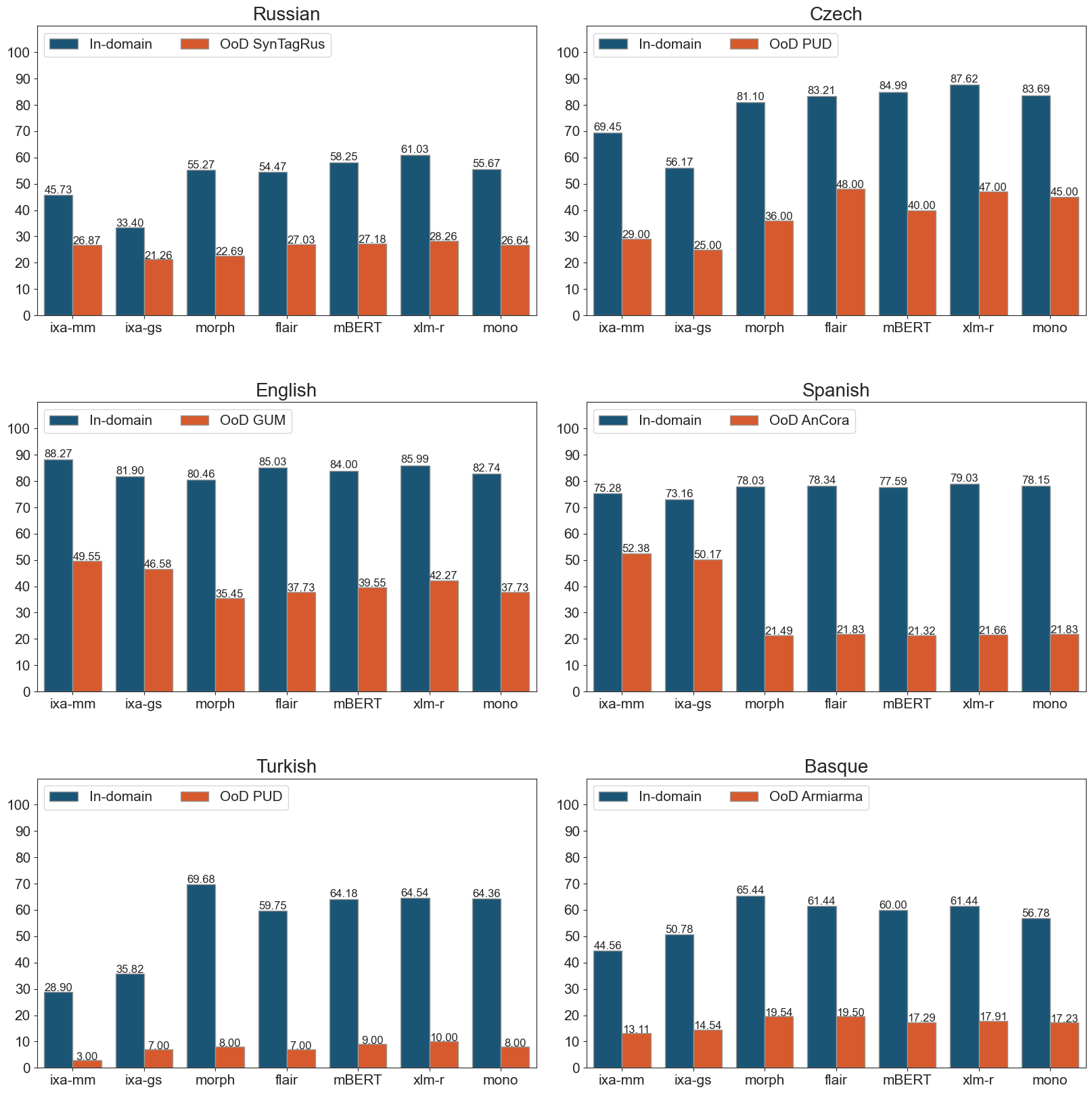}
	\caption{Sentence accuracy results for in- and out-of-domain settings.}
	\label{fig:sentence_accuracy.png}
\end{figure}

Figure \ref{fig:sentence_accuracy.png} reports the sentence accuracy of the six
languages we used in our experiments both for in- and out-of-domain. In
contrast to the word accuracies reported in Figure \ref{fig:inD_outD_overall.png}, we can see that the corresponding
sentence accuracy results drop significantly. In addition to demonstrating that
lemmatizers have a large margin of improvement, sentence accuracy allows us to
better discriminate between different models. We can see this
phenomenon in the English and Spanish results. Thus, while every model obtained
very similar in-domain word accuracy in Spanish, using sentence accuracy helps
to discriminate between the statistical and the neural
lemmatizers. Furthermore, it also shows that among the neural models XLM-RoBERTa clearly
outperforms the rest of the models by almost 1 percent.

The effect of sentence accuracy for the in-domain evaluation is vastly
magnified when considering out-of-domain performance, with the extremely low scores
across languages providing further evidence of how far lemmatization remains from being
solved. 

\subsection{Analyzing word accuracy per SES}\label{sec:macro-accuracy-ses}

The next natural step in our analysis is identifying which specific cases are
most difficult for lemmatizers. In order to do so, we look at the
word accuracy for each of the SES labels automatically induced from the data. In order to
illustrate this point, we took XLM-RoBERTa-base as an example use case
and analyze their predictions for the languages which could be inspected in-house,
namely, Basque, English, Spanish and Russian. Thus, Table
\ref{tab:ses-freq-wacc} presents examples and results for the 10 most frequent
SES for each of these 4 languages development sets.

\begin{table}
    \footnotesize
     \centering
    \begin{tabularx}{\textwidth}{c|l|l|l|c|c|l}
      \toprule
       & SES & Casing & Edit script & W.acc & \% & Examples \\ \midrule
       &↓0;d¦+ & all low & do nothing & 99.29 & 76.87\% & positive$\rightarrow$\emph{positive} \\
       &↑0¦↓1;d¦+ & 1st up & do nothing & 96.29 & 6.97\% & Martin$\rightarrow$\emph{Martin}\\
       &↓0;d¦-+  & all low & remove last ch & 98.58 & 5.52\% & things$\rightarrow$\emph{thing}\\ 
       &↓0;abe & all low & ignore form, use \emph{be} & 99.81 & 2.02\% & is$\rightarrow$\emph{be} \\
        en &↓0;d¦--+ & all low & remove 2 last ch & 97.42 & 1.52\% & does$\rightarrow$\emph{do} \\
        &↓0;d¦---+ & all low & remove 3 last ch & 96.45 & 1.10\% & trying$\rightarrow$\emph{try} \\
        &↑0¦↓-1;d¦+ & all up & do nothing &  94.22 & 0.68\% & NASA$\rightarrow$\emph{NASA} \\
        &↓0;d--+b¦+& all low & first 2 char to \emph{b} & 99.33 & 0.59\% & are$\rightarrow$\emph{be} \\
        &↓0;d¦-+v+e+ & all low & last ch to \emph{ve} & 100.00 & 0.51\% & has$\rightarrow$\emph{have}\\
        &↓0;d¦---+e+ & all low & 3 last ch to \emph{e} & 96.23 & 0.42\% & driving$\rightarrow$\emph{drive} \\ \midrule
        
        &↓0;d¦+ & all low & do nothing &  99.36 & 72.40\% & acuerdo$\rightarrow$\emph{acuerdo} \\
        &↓0;d¦-+ & all low & del last ch & 97.22 & 5.29\% & estrellass$\rightarrow$\emph{estrella}  \\
        &↓0;d+e¦-+ & all low & add \emph{e}, del last ch &  96.73 & 3.36\% & la$\rightarrow$\emph{el} \\
        &↓0;d¦-+o+ & all low & del last ch, add \emph{o} & 96.21 & 2.37\% & una$\rightarrow$\emph{uno} \\
        es & ↓0;d+e¦--+ & all low & add \emph{e}, del 2 last ch &  99.78 & 2.13\% & los$\rightarrow$\emph{el} \\
        &↓0;d¦--+ & all low & del 2 last ch & 97.36 & 1.40\% & flores$\rightarrow$\emph{flor} \\
        &↓0;aél & all low & ignore form, use \emph{él} & 99.83 & 1.32\% & se$\rightarrow$\emph{él} \\
        &↓0;d¦+r+ & all low & add \emph{r} & 100.00 & 0.91\% & hace$\rightarrow$\emph{hacer}\\
        &↓0;d¦+o+ & all low &  add \emph{o} & 97.73 & 0.91\% & primer$\rightarrow$\emph{primero}\\
        &↓0;d¦-+a+r+ & all low & del last ch, add \emph{ar} & 98.07 & 0.83\% & desarrolló$\rightarrow$\emph{desarollar} \\ \midrule
        
        &↓0;d¦+ & all low & do nothing & 99.16 & 57.80\% & Петербург$\rightarrow$\emph{Петербург} \\ 
        &↓0;d¦-+ & all low & del last ch & 97.67 & 6.97\% & церковью$\rightarrow$\emph{церковь} \\
        &↓0;d¦-+а+ & all low & del last ch, add \emph{а} & 96.65 & 3.32\% & экономику$\rightarrow$\emph{экономика} \\
        &↓0;d¦-+й+ & all low & del last ch, add \emph{й} & 96.08 & 3.10\% & городское$\rightarrow$\emph{городской}\\
        ru & ↓0;d¦--+ & all low & del 2 last ch & 99.03 & 2.10\% & странами$\rightarrow$\emph{страна} \\
        &↓0;d¦-+е+ & all low & del last ch, add \emph{е} & 98.04 & 2.07\% & моря$\rightarrow$\emph{море} \\
        &↓0;d¦-+я+ & all low & del last ch, add \emph{я} & 97.83 & 1.86\% & историю$\rightarrow$\emph{история}\\
        &↓0;d¦-+т+ь+ & all low & del last ch, add \emph{ть} & 98.88 & 1.81\% & получил$\rightarrow$\emph{получить} \\
        &↓0;d¦-+ь+ & all low & del last ch, add \emph{ь} & 93.94 & 1.67\% & сентября$\rightarrow$\emph{сентябрь} \\
        &↓0;d¦--+т+ь+ & all low & del 2 last, add \emph{ть} & 98.10 & 1.60\% & были$\rightarrow$\emph{быть} \\ \midrule
        
        &↓0;d¦+ & all low & do nothing & 99.05  & 49.63\% & sartu$\rightarrow$\emph{sartu} \\
        &↓0;d¦--+ & all low & remove 2 last ch & 97.72 & 9.93\% & librean$\rightarrow$\emph{libre} \\
        &↓0;d¦-+ & all low & remove last ch & 96.27 & 6.54\% & korrikan$\rightarrow$\emph{korrika} \\
        &↓0;d¦---+ & all low & remove 3 last ch & 93.24 & 3.60\% & aldaketarik$\rightarrow$\emph{aldaketa} \\
        eu &↑0¦↓1;d¦+ & 1st up & do nothing & 98.54 & 3.46\%\ & MAPEI$\rightarrow$\emph{Mapei} \\
        &↓0;d¦----+ & all low & del 4 last ch & 93.00 & 2.52\% & lagunaren$\rightarrow$\emph{lagun}\\
        &↑0¦↓1;d¦--+ & 1st up & del 2 last ch & 95.54 & 1.88\% & Egiptora$\rightarrow$\emph{Egipto} \\
        &↓0;d-+i+z & all low & del 1st ch,  & 100.00 & 1.38\% & da$\rightarrow$\emph{izan} \\
        &¦+n+ & &add \emph{iz},\emph{n} & & & \\
        &↑0¦↓1;d¦-+ & 1st up & del last ch & 90.08 & 1.10\% & Frantziak$\rightarrow$\emph{Frantzia} \\ \bottomrule
    \end{tabularx}
    \caption {10 most frequent SES, brief description, corresponding word accuracy, weight (in \%) in the corpus and examples of words and their lemmas for English, Spanish, Russian and Basque; SES are computed following UDPipe's method \citep{straka-etal-2019-udpipe}.}.
    \label{tab:ses-freq-wacc}
\end{table}

As we can see in Table \ref{tab:ses-freq-wacc}, the most common lemma
transformation to be learned is based on the edit script ``do nothing'',
namely, the lemmatizer needs to learn that the lemma and the word have the same
form. It is also 
interesting to see how the ratio of such lemma type changes across languages,
from English, where such cases are observed in almost 77\% of the cases to
Basque, where only half of the lemmas correspond to this rule. However, in
terms of word accuracy, the results are remarkably similar for all 4 languages,
in the range of 99-99.30\%. This demonstrates that the traditional evaluation
method greatly overestimates the lemmatizers' performance.

By looking at other specific cases, we can see that in English problematic
examples to learn are those related to the casing of some characters (e.g. Martin $\rightarrow$
\emph{Martin}, NASA $\rightarrow$ \emph{NASA}). Other noticeable issue refers to
the verbs in gerund form (e.g. trying $\rightarrow$ \emph{try},
driving $\rightarrow$ \emph{drive}).

With respect to Spanish interesting difficult lemmas are observed with articles
in feminine form (e.g. la $\rightarrow$ \emph{el}, una $\rightarrow$
\emph{uno}), where the masculine form is considered the canonical form or
lemma, and feminine articles and adjectives should be lemmatized by changing
the gender of the word from female to male.

In Russian the most challenging case corresponds to the lemmatization of the
nouns that end with a soft sign {ь} with the word accuracy for this SES
as low as 93.94\%. The possible reason of such low accuracy could be the absence of a
specific grammar rule that defines the gender of such nouns and, therefore, the
termination these nouns have in different cases. The second lowest accuracy
among the 10 most popular SES in Russian is for adjectives, cases in which to
obtain the lemma one should delete the last character of the word and add a letter
й (pronounced as \emph{iy kratkoe}, short y), that in Russian
determines the suffix for some masculine nouns and adjectives in singular and nominative case. The words
could be in different cases and genders, so it is necessary to know such
information for correct lemmatization (e.g. городско\textbf{е} $\rightarrow$
\emph{городско\textbf{й}} (neutral gender, nominative case), семейны\textbf{м}
$\rightarrow$ \emph{семейны\textbf{й}} (masculine gender, instrumental case)).

Finally, for Basque the most problematic cases with a rather low word accuracy
of only 90.08\% can be found among the nouns in ergative (e.g.
Frantzia\textbf{k} $\rightarrow$ \emph{Frantzia}) or locative cases (e.g.
Mosku\textbf{n} (in Moscow) $\rightarrow$ \emph{Mosku},
Katalunia\textbf{n} (in Catalonia) $\rightarrow$ \emph{Katalunia}). The other two
most difficult SES occur when the word forms are in possessive case (e.g.
lagun\textbf{aren} $\rightarrow$ \emph{lagun}) and for nouns in indefinite form 
(e.g., aldaketa\textbf{rik} (change) $\rightarrow$ \emph{aldaketa}). 

It should be noted that an extra obstacle to improving some of
these difficult cases is the low number of samples available. Nonetheless, this
analysis shows that lemmatizers still do not properly learn to lemmatize
relatively common word forms.

\subsection{Generalization Capabilities of Language Models}

In this subsection we aim to analyze the generalization capabilities of a MLM
such as XLM-RoBERTa-base in the lemmatization task. More specifically, we will
discuss two issues: (i) whether MLMs simply memorize the SES lemma classes
during fine-tuning and (ii) whether the good performance of MLMs in this task
might be due to some test data
contamination.\footnote{\url{https://hitz-zentroa.github.io/lm-contamination/}}

In order to address the first point, we evaluate the performance of
XLM-RoBERTa-base, fine-tuned without morphological features, for those words seen
during fine-tuning (in-vocabulary words) with respect to out-of-vocabulary
occurrences.

\begin{table}[h]
	\small
     \centering
    \newcolumntype{Y}{>{\centering\arraybackslash}X}
    \begin{tabularx}{\textwidth}{Y|YY|YY}
    \toprule
    & \multicolumn{2}{c}{in-domain} & \multicolumn{2}{c}{out-of-domain}\\ \midrule
    & in-vocabulary & out-of-vocabulary & in-vocabulary & out-of-vocabulary \\ \midrule
    en & 97.25 & 90.55 & 92.56 & 81.11 \\
    es & 98.06 & 93.54 & 82.53 & 60.07 \\
    eu & 96.65 & 82.85 & 87.92 & 71.08 \\
    ru & 98.62 & 90.23 & 89.19 & 77.50 \\
    cz & 99.21 & 93.28 & 98.08 & 88.66 \\
    tr & 97.84 & 84.54 & 92.34 & 68.39 \\
     \bottomrule
     \end{tabularx}
    \caption{Averaged word accuracy for in-vocabulary and out-of-vocabulary words for XLM-RoBERTa-base model (original setting). Corpora: English - EWT (in-domain), GUM (out-of-domain); Spanish - GSD (in-domain), AnCora (out-of-domain); Basque - BDT (in-domain), Armiarma (out-of-domain); Russian - GSD (in-domain), SynTagRus (out-of-domain); Czech - CAC (in-domain), PUD (out-of-domain); Turkish - IMST (in-domain), PUD (out-of-domain).}
    \label{tab:in-vocab-original}
    \end{table}  

\begin{table}[h]
	\small
     \centering
    \newcolumntype{Y}{>{\centering\arraybackslash}X}
    \begin{tabularx}{\textwidth}{Y|YY|YY}
    \toprule
    & \multicolumn{2}{c|}{in-domain} & \multicolumn{2}{c}{out-of-domain}\\ \midrule
    & in-vocabulary & out-of-vocabulary & in-vocabulary & out-of-vocabulary \\ \midrule
    en & 96.48 & 88.95 & 89.84 & 75.94 \\
    es & 98.54 & 93.11 & 81.71 & 47.10 \\
    ru & 98.70 & 92.28 & 89.22 & 59.13 \\
    cz & 96.26 & 83.51 & 90.42 & 82.35 \\
    tr & 90.11 & 70.15 & 88.63 & 58.79 \\ \bottomrule
     \end{tabularx}
    \caption{Averaged word accuracy for in-vocabulary and out-of-vocabulary words for XLM-RoBERTa-base model (reversed setting). Corpora: English - GUM (in-domain), EWT (out-of-domain); Spanish - AnCora (in-domain), GSD (out-of-domain); Russian - SynTagRus (in-domain), GSD (out-of-domain); Czech - PUD (in-domain), CAC (out-of-domain); Turkish - PUD (in-domain), IMST (out-of-domain).}
    \label{tab:in-vocab-reversed}
    \end{table}

Tables \ref{tab:in-vocab-original} and \ref{tab:in-vocab-reversed} report the
results for both original and reversed settings and in- and out-of-domain
evaluations. It is noticeable that the model performs very well on
out-of-vocabulary words, also in the out-of-domain evaluation, which would
seem to indicate that XLM-RoBERTa is generalizing beyond the words seen during training. This seems to
be confirmed also by looking at the Spanish and Russian results. It should be
remembered that, while in the reversed setting the training data for Spanish
(AnCora, $~$500K tokens) and Russian (SynTagRus, $~$900K words) is much larger
than in the original setting (both GSD), the obtained results reflect roughly the same trend.

Finally, we should consider whether a MLM such as XLM-RoBERTa has
already seen the datasets we are experimenting with during pre-training,
namely, whether XLM-RoBERTa has been
contaminated.\footnote{\url{https://hitz-zentroa.github.io/lm-contamination/blog/}} First,
it should be noted that CC-100, the corpus used to generate XLM-RoBERTa, was
constructed by processing the CommonCrawl snapshots from between January and
December 2018. Second, the SIGMORPHON data we are using was released in
2019\footnote{First GitHub commit December 19, 2018.} with the test data
including gold standard lemma and UniMorph annotations being released in April 2019.
Third and most importantly, XLM-RoBERTa does not see the lemmas themselves
during training or inference, but the SES classes we automatically generate in
an ad-hoc manner for the experimentation. The datasets containing both the
words and the SES classes used have not been yet made publicly available.

Based on this, it is possible to say that XLM-RoBERTa seems to generalize over
unseen words and that its performance is not justified by any form of language
model contamination.

\subsection{Analyzing Spanish out-of-domain results}

In Section \ref{sec:out-of-domain} we saw that out-of-domain performance of transformer-based
models for Spanish was not following the pattern of the rest of the languages.
Instead, they were 6-7\% worse than the results obtained by the IXA pipes
statistical lemmatizers (ixa-pipe-mm and ixa-pipe-gs). By
checking the most common error patterns of XLM-RoBERTa-base, we found out that most
of the performance loss was caused by inconsistencies in the manual annotation
of lemmas between the data used for in-domain and out-of-domain evaluation. More specifically, the GSD Spanish
corpus included in UniMorph wrongly annotates lemmas for proper names
such as Madrid, London or Paris entirely in lowercase, namely, \emph{madrid},
\emph{london} and \emph{paris}. However, the AnCora Spanish corpus used for
out-of-domain evaluation correctly annotates these cases specifying their
corresponding lemmas with the first character in uppercase. This inconsistency
results in 3781 examples of proper names in the AnCora test set which are all
lemmatized following the pattern seen during training with the GSD training
set. Consequently, the word accuracy obtained by the model for this type of examples in the AnCora
test set is 0\%. In order to confirm this issue, we corrected the wrongly
annotated proper names in the GSD training data, fine-tuned again the model and
saw the out-of-domain performance of XLM-RoBERTa-base go up from 90.26\% to
96.75\%, a more consistent result with respect to the out-of-domain
scores for the other 5 languages.

This issue manifests the importance of consistent manual annotation across
corpora from different domains in order to fairly evaluate out-of-domain
performance of contextual lemmatizers.

\section{Concluding Remarks}\label{sec:conclusion}

Lemmatization remains an important natural language processing task, especially
for languages with high-inflected morphology. In this paper we provide an
in-depth study on the role of morphological information to learn contextual
lemmatizers. By taking a language sample of varied morphological complexity, we
have analyzed whether fine-grained morphological signal is indeed beneficial
for contextual lemmatization. Furthermore, and in contrast to previous work, 
we also evaluate lemmatizers in an out-of-domain setting, which
constitutes, after all, their most common application use. Our results
empirically demonstrate that informing lemmatizers with fine-grained
morphological features during training is not that beneficial, not even for
agglutinative languages. In fact, modern contextual word representations
seem to implicitly encode enough morphological information to obtain good
contextual lemmatizers without seeing any explicit morphological signal.
Finally, good out-of-domain performance can be achieved using simple UPOS tags
or without any explicit morphological signal.

Therefore, our results suggest that an optimal solution among all the
options considered would be to develop lemmatizers by fine-tuning a large MLM
such as XLM-RoBERTa-large without any explicit morphological signal. Addressing
lemmatization as a token classification task results in highly
competitive and robust lemmatizers with results over or close to the
state-of-the-art obtained with more complex methods
\citep{straka-etal-2019-udpipe}.

Furthermore, we have discussed current evaluation practices for lemmatization,
showing that using simple word accuracy is not adequate to clearly discriminate
between models, as it provides a deceptive view regarding the performance of
lemmatizers. An additional analysis looking at specific lemma classes (SES)
has shown that many common word forms are still not properly predicted. The
conclusion is that lemmatization remains a challenging task. Future work is
therefore needed to improve out-of-domain results. Furthermore, it is perhaps a
good time to propose an alternative word-level metric to evaluate lemmatization
that, complemented with sentence accuracy, may provide a more realistic view of
the performance of contextual lemmatizers.

\section*{Acknowledgements}

This work has been supported by the HiTZ center and the Basque Government
(Research group funding IT-1805-22). Olia Toporkov is funded by a UPV/EHU grant
``Formación de Personal Investigador''. We also thank the funding from
the following MCIN/AEI/10.13039/501100011033 projects: (i) DeepKnowledge
(PID2021-127777OB-C21) and ERDF A way of making Europe; (ii) Disargue
(TED2021-130810B-C21) and European Union NextGenerationEU/PRTR (iii) Antidote
(PCI2020-120717-2) and European Union NextGenerationEU/PRTR. 
Rodrigo Agerri currently holds the RYC-2017-23647 (MCIN/AEI/10.13039/501100011033
and ESF Investing in your future) fellowship.

\bibliography{bibliography,anthology}

\clearpage

\appendix

\section{Detailed Lemmatization Results}\label{ap:appendixA}

\begin{table}[h]
\small
    \centering
    \newcolumntype{Y}{>{\centering\arraybackslash}X}
    \begin{tabularx}{\textwidth}{Y|YYYYYYYYY}
    \toprule
    & ixa-mm & ixa-gs & morph & flair & mBERT & xlm-r & mono & base & UDPipe \\ \midrule
    en & \underline{99.06} & 98.95 & 98.10 & 98.58 & 98.56 & 98.76 & 98.49 & 97.68 & 99.01 \\
    es & 98.89 & 98.74 & 99.02 & 99.02 & 99.01 & \underline{99.08} & 99.04 & 98.42 & 99.31\\
    ru & 95.07 & 93.22 & 96.30 & 96.18 & 96.70 & \underline{97.08} & 96.55 & 95.67 & 97.77\\
    eu & 93.41 & 94.06 & \underline{96.39} & 96.09 & 95.71 & 95.98 & 95.51 & 96.07 & 97.14 \\
    cz & 97.86 & 96.63 & 98.76 & 98.87 & 99.07 & \underline{99.25} & 99.01 & 97.82 & 99.31 \\
    tr & 85.52 & 86.57 & \underline{96.18} & 93.98 & 95.15 & 95.38 & 95.20 & 96.41 & 96.84 \\
     \bottomrule 
     \end{tabularx}
    \caption{Overall in-domain lemmatization results for models trained with and without explicit morphological features; monolingual  transformers: Russian - ruBERT, Czech - slavicBERT, Basque -BERTeus, Turkish - BERTurk, English - RoBERTa, Spanish - BETO.}
    \label{tab:Results on the test sets of the systems that use morphology to train contextual lemmatizers in-domain (original setting).}
    \end{table}

\begin{table}[h]
\small
     \centering
    \newcolumntype{Y}{>{\centering\arraybackslash}X}
    \begin{tabularx}{\textwidth}{Y|YYYYYYY}
    \toprule
    & ixa-mm & ixa-gs & morph & flair & mBERT & xlm-r & mono\\ \midrule
    en & \underline{95.16} & 95.13 & 92.92 & 93.42 & 93.50 & 93.56 & 93.39 \\
    es & \underline{97.59} & 97.45 & 90.35 & 90.29 & 90.27 & 90.26 & 90.34 \\
    ru & \underline{91.00} & 88.66 & 87.57 & 89.90 & 90.07 & 90.53 & 89.71 \\
    eu & 85.33 & 86.31 & \underline{89.03} & 88.76 & 87.79 & 88.15 & 87.62 \\
    cz & 92.33 & 91.81 & 91.92 & 95.02 & 94.72 & \underline{95.18} & 94.40\\
    tr & 80.33 & 80.50 & 84.74 & 83.51 & 84.40 & \underline{84.90} & 84.46 \\
     \bottomrule
     \end{tabularx}
    \caption{Overall out-of-domain lemmatization results for models with and without explicit morphological features; monolingual  transformers: Russian - ruBERT, Czech - slavicBERT, Basque -BERTeus, Turkish - BERTurk, English - RoBERTa, Spanish - BETO.}
    \label{tab:Results on the test sets of the systems that use morphology to train contextual lemmatizers out-of-domain (original setting).}
    \end{table}  

\begin{table}[h]
\small
     \centering
    \newcolumntype{Y}{>{\centering\arraybackslash}X}
    \begin{tabularx}{\textwidth}{Y|YYYYYYY}
    \toprule
    & ixa-mm & ixa-gs & morph & flair & mBERT & xlm-r & mono\\ \midrule
    en & \underline{88.27} & 81.90 & 80.46 & 85.03 & 84.00 & 85.99 & 82.74 \\
    es & 75.28 & 73.16 & 78.03 & 78.34 & 77.59 & \underline{79.03} & 78.15 \\
    ru & 45.73 & 33.40 & 55.27 & 54.47 & 58.25 & \underline{61.03} & 55.67 \\
    eu & 44.56 & 50.78 & \underline{65.44} & 61.44 & 60.00 & 61.44 & 56.78 \\
    cz & 69.45 & 56.17 & 81.10 & 83.21 & 84.99 & \underline{87.62} & 83.69 \\
    tr & 28.90 & 35.82 & \underline{69.68} & 59.75 & 64.18 & 64.54 & 64.36 \\
     \bottomrule
     \end{tabularx}
     \caption{In-domain sentence accuracy results; monolingual  transformers: Russian - ruBERT, Czech - slavicBERT, Basque -BERTeus, Turkish - BERTurk, English - RoBERTa, Spanish - BETO.}
    \label{tab:Sentence accuracy results in-domain (original setting).}
    \end{table}

\begin{table}[h]
\small
     \centering
    \newcolumntype{Y}{>{\centering\arraybackslash}X}
    \begin{tabularx}{\textwidth}{Y|YYYYYYY}
    \toprule
    & ixa-mm & ixa-gs & morph & flair & mBERT & xlm-r & mono\\ \midrule
    en & \underline{49.55} & 46.58 & 35.45 & 37.73 & 39.55 & 42.27 & 37.73 \\
    es & \underline{52.38} & 50.17 & 21.49 & 21.83 & 21.32 & 21.66 & 21.83 \\
    ru & 26.87 & 21.26 & 22.69 & 27.03 & 27.18 & \underline{28.26} & 26.64 \\
    eu & 13.11 & 14.54 & \underline{19.54} & 19.50 & 17.29 & 17.91 & 17.23 \\
    cz & 29.00 & 25.00 & 36.00 & \underline{48.00} & 40.00 & 47.00 & 45.00 \\
    tr & 3.00 & 7.00 & 8.00 & 7.00 & 9.00 & \underline{10.00} & 8.00 \\
     \midrule
     \end{tabularx}
    \caption{Out-of-domain sentence accuracy results; monolingual  transformers: Russian - ruBERT, Czech - slavicBERT, Basque -BERTeus, Turkish - BERTurk, English - RoBERTa, Spanish - BETO.}
    \label{tab:Sentence accuracy results out-of-domain (original setting).}
    \end{table}  


\begin{table}[h]
\small
     \centering
    \newcolumntype{Y}{>{\centering\arraybackslash}X}
    \begin{tabularx}{\textwidth}{Y|YYYYYYYY}
    \toprule
    & ixa-mm & ixa-gs & morph & mBERT & xlm-r & mono & base & UDPipe \\ \midrule
    en & 97.56 & 97.12 & \underline{97.78} & 97.19 & 97.70 & 96.90 & 97.41 & 98.63 \\
    es & 98.70 & 98.53 & 98.98 & 99.14 & 99.19 & \underline{99.23} & 98.54 & 99.46 \\
    ru & 96.76 & 96.84 & 96.93 & 98.66 & \underline{98.93} & 98.67 & 95.92 & 98.92\\
    cz & 89.59 & 88.03 & 93.11 & 93.01 & 93.06 & \underline{93.37} & 93.58 & 98.13 \\
    tr & 77.77 & 78.33 & \underline{87.02} & 83.40 & 85.07 & 82.56 & 86.02 & 89.03 \\
     \midrule
     \end{tabularx}
    \caption{Overall in-domain lemmatization results  (reversed setting) for models with and without explicit morphological features; monolingual  transformers: Russian - ruBERT, Czech - slavicBERT, Basque -BERTeus, Turkish - BERTurk, English - RoBERTa, Spanish - BETO.}
    \label{tab:Results on the test sets of the systems that use morphology to train contextual lemmatizers in-domain (reversed setting).}
    \end{table}

\begin{table}[h]
\small
     \centering
    \newcolumntype{Y}{>{\centering\arraybackslash}X}
    \begin{tabularx}{\textwidth}{Y|YYYYYY}
    \toprule
    & ixa-mm & ixa-gs & morph & mBERT & xlm-r & mono\\ \midrule
    en & \underline{91.22} & 90.55 & 88.97 & 90.80 & 91.21 & 90.94\\
    es & \underline{87.90} & 87.47 & 87.50 & 87.51 & 87.65 & 87.33 \\
    ru & 85.37 & 86.25 & 86.10 & 87.51 & \underline{88.43} & 87.64 \\
    cz & 86.09 & 83.94 & 89.13 & 89.73 & \underline{90.10} & 89.17\\
    tr & 70.61 & 70.95 & \underline{81.03} & 77.01 & 78.22 & 76.87\\
     \bottomrule 
     \end{tabularx}
    \caption{Overall out-of-domain lemmatization results (reversed setting) for models with and without explicit morphological features; monolingual  transformers: Russian - ruBERT, Czech - slavicBERT, Basque -BERTeus, Turkish - BERTurk, English - RoBERTa, Spanish - BETO.}
    \label{tab:Results on the test sets of the systems that use morphology to train contextual lemmatizers out-of-domain (reversed setting).}
    \end{table} 

\end{document}